\pdfoutput=1
\documentclass[11pt]{article}

\usepackage[]{acl}

\usepackage{times}
\usepackage{latexsym}

\usepackage[T1]{fontenc}
\usepackage[utf8]{inputenc}

\usepackage{microtype}

\usepackage{algorithm}
\usepackage{algorithmic}
\usepackage{amsthm,amsmath,amssymb}
\usepackage{multirow}
\usepackage{booktabs}
\usepackage{graphicx}
\usepackage{url}

\title{Learning to Win Lottery Tickets in BERT Transfer via Task-agnostic Mask Training}

\def\first{$^1$}
\def\second{$^2$}
\def\third{$^3$}
\def\comma{$^,$}
\def\star{$^*$}

\author{Yuanxin Liu\first\comma\second 
, Fandong Meng\third
, Zheng Lin\first\comma\second\star, \\
  \textbf{Peng Fu}\first\star
, \textbf{Yanan Cao}\first\comma\second
, \textbf{Weiping Wang}\first
, \textbf{Jie Zhou}\third
\\
{\first {Institute of Information Engineering, Chinese Academy of Sciences, Beijing, China}}  \\
{\second {School of Cyber Security, University of Chinese Academy of Sciences, Beijing, China}} \\
{\third {Pattern Recognition Center, WeChat AI, Tencent Inc, China}} \\
{\tt \small{\{liuyuanxin,linzheng,fupeng,caoyanan,wangweiping\}@iie.ac.cn}}, \\
{\tt \small{ \{fandongmeng,withtomzhou\}@tencent.com}}}

\begin{document}
\maketitle

\newcommand\blfootnote[1]{% 
\begingroup 
\renewcommand\thefootnote{}\footnote{#1}% 
\addtocounter{footnote}{-1}% 
\endgroup
}
\blfootnote{Joint work with Pattern Recognition Center, WeChat AI, Tencent Inc, China.\star{Zheng Lin and Peng Fu are the corresponding authors.}}

\begin{abstract}
Recent studies on the \textit{lottery ticket hypothesis} (LTH) show that pre-trained language models (PLMs) like BERT contain \textit{matching subnetworks} that have similar transfer learning performance as the original PLM. These subnetworks are found using magnitude-based pruning. In this paper, we find that the BERT subnetworks have even more potential than these studies have shown. Firstly, we discover that the success of magnitude pruning can be attributed to the preserved pre-training performance, which correlates with the downstream transferability. Inspired by this, we propose to directly optimize the subnetwork structure towards the pre-training objectives, which can better preserve the pre-training performance. Specifically, we train binary masks over model weights on the pre-training tasks, with the aim of preserving the universal transferability of the subnetwork, which is agnostic to any specific downstream tasks. We then fine-tune the subnetworks on the GLUE benchmark and the SQuAD dataset. The results show that, compared with magnitude pruning, mask training can effectively find BERT subnetworks with improved overall performance on downstream tasks. Moreover, our method is also more efficient in searching subnetworks and more advantageous when fine-tuning within a certain range of data scarcity. Our code is available at \url{https://github.com/llyx97/TAMT}.
\end{abstract}

\section{Introduction}
\label{sec:intro}
The NLP community has witnessed a remarkable success of pre-trained language models (PLMs). After being pre-trained on unlabelled corpus in a self-supervised manner, PLMs like BERT \cite{BERT} can be fine-tuned as a universal text encoder on a wide range of downstream tasks. However, the growing performance of BERT is driven, to a large extent, by scaling up the model size, which hinders the fine-tuning and deployment of BERT in resource-constrained scenarios. 

At the same time, the \textit{lottery ticket hypothesis} (LTH) \cite{LTH} emerges as an active sub-field of model compression. The LTH states that randomly initialized dense networks contain sparse \textit{matching subnetworks}, i.e., winning tickets (WTs), that can be trained in isolation to similar test accuracy as the full model. The original work of LTH and subsequent studies have demonstrated that such WTs do exist at random initialization or an early point of training \cite{LTAtScale,LinearMode}. This implicates that it is possible to reduce training and inference cost via LTH.

%%%%%%%%%%%%%%%%%%%%%%%%%%%%%%%%%%%%%%%%%%%%%%
\begin{figure}[t]
\centering
\includegraphics[width=1.0\linewidth]{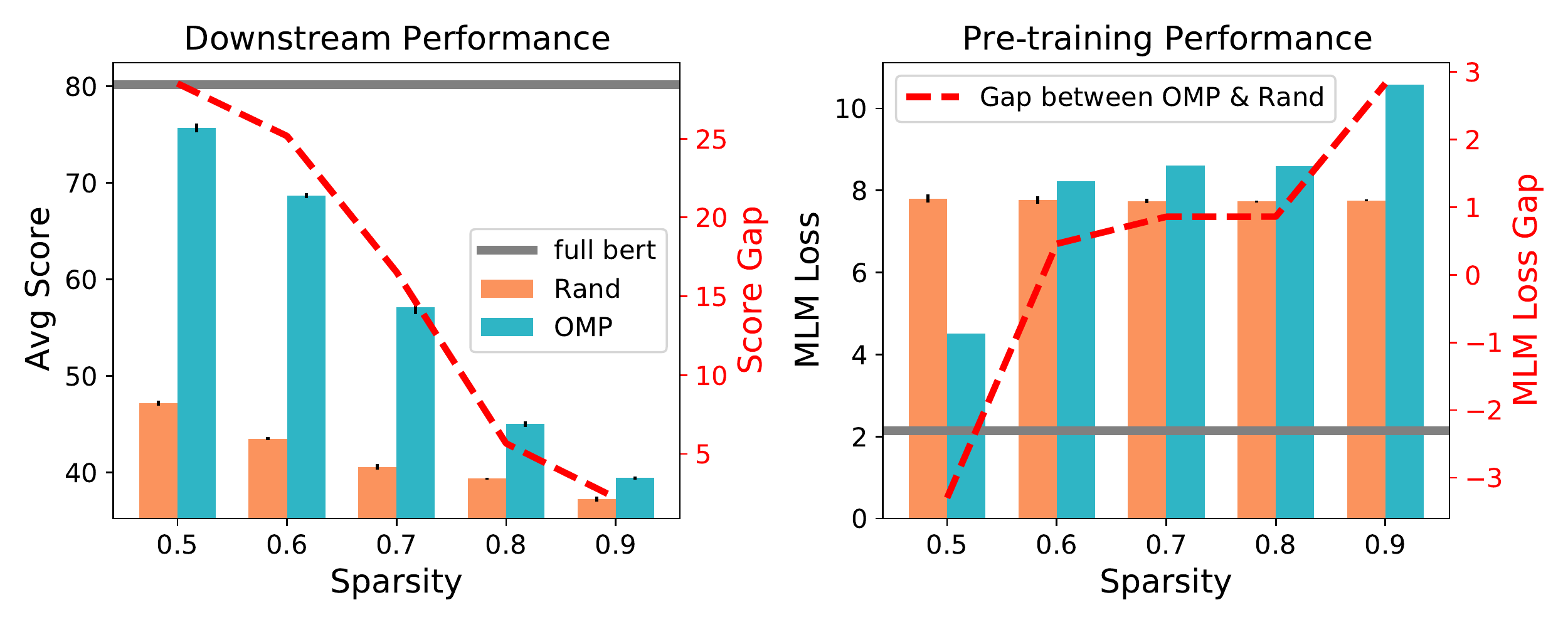}
\caption{Average downstream performance (left) and pre-training performance (right) of OMP and random subnetworks of $\mathrm{BERT}_{\mathrm{BASE}}$. See Appendix \ref{sec:appendix_a1} for the downstream results of each task.}
\label{fig:motivation}
\end{figure}
%%%%%%%%%%%%%%%%%%%%%%%%%%%%%%%%%%%%%%%%%%%%%%

Recently, \citet{BERT-LT} extend the original LTH to the \textit{pre-training and fine-tuning} paradigm, exploring the existence of matching subnetworks in pre-trained BERT. Such subnetworks are smaller in size, while they can preserve the universal transferability of the full model. Encouragingly, \citet{BERT-LT} demonstrate that BERT indeed contains matching subnetworks that are transferable to multiple downstream tasks without compromising accuracy. These subnetworks are found using iterative magnitude pruning (IMP) \cite{HanNips} on the pre-training task of masked language modeling (MLM), or by directly compressing BERT with oneshot magnitude pruning (OMP), both of which are agnostic to any specific task.

In this paper, we follow \citet{BERT-LT} to study the question of LTH in BERT transfer learning. We find that there is a correlation, to certain extent, between the performance of a BERT subnetwork on the pre-training task (right after pruning), and its downstream performance (after fine-tuning). As shown by Fig. \ref{fig:motivation}, the OMP subnetworks significantly outperform random subnetworks at 50\% sparsity in terms of both MLM loss and downstream score. However, with the increase of model sparsity, the downstream performance and pre-training performance degrade simultaneously. This phenomenon suggests that we might be able to further improve the transferability of BERT subnetworks by discovering the structures that better preserve the pre-training performance.

To this end, we propose to search transferable BERT subnetworks via \textbf{T}ask-\textbf{A}gnostic \textbf{M}ask \textbf{T}raining (TAMT), which learns selective binary masks over the model weights on pre-training tasks. In this way, the structure of a subnetwork is directly optimized towards the pre-training objectives, which can preserve the pre-training performance better than heuristically retaining the weights with large magnitudes. The training objective of the masks is a free choice, which can be designed as any loss functions that are agnostic to the downstream tasks. In particular, we investigate the use of MLM loss and a loss based on knowledge distillation (KD) \cite{KD}.

To examine the effectiveness of the proposal, we train the masks on the WikiText dataset \cite{WikiText} for language modeling and then fine-tune the searched subnetworks on a wide variety of downstream tasks, including the GLUE benchmark \cite{GLUE} for natural language understanding (NLU) and the SQuAD dataset \cite{SQuAD} for question answering (QA). The empirical results show that, through mask training, we can indeed find subnetworks with lower pre-training loss and better downstream transferability than OMP and IMP. Compared with IMP, which also involves training (the weights) on the pre-training task, mask training requires much fewer training iterations to reach the same performance. Moreover, the subnetworks found by mask training are generally more robust when being fine-tuned with reduced data, as long as the training data is not extremely scarce.

In summary, our contributions are:
\begin{itemize}
\item We find that the pre-training performance of a BERT subnetwork correlates with its downstream transferability, which provides a useful insight for the design of methods to search transferable BERT subnetworks.
\item Based on the above finding, we propose to search subnetworks by learning binary masks over the weights of BERT, which can directly optimize the subnetwork structure towards the given pre-training objective.
\item Experiments on a variety of NLP tasks show that subnetworks found by mask training have better downstream performance than magnitude pruning. This suggests that BERT subnetworks have more potential, in terms of universal downstream transferability, than existing work has shown, which can facilitate our understanding and application of LTH on BERT.
\end{itemize}

\section{Related Work}
\subsection{The Lottery Ticket Hypothesis}
The lottery ticket hypothesis \cite{LTH} suggests the existence of matching subnetworks, at random initialization, that can be trained in isolation to reach the performance of the original network. However, the matching subnetworks are found using IMP, which typically requires more training cost than the full network. There are two remedies to overcome this problem: \citet{OneTicket} proposed to transfer the WT structure from source tasks to related target tasks, so that no further searching is required for the target tasks. \citet{Early-Bird} draw \textit{early-bird tickets} (prune the original network) at an early stage of training, and only train the subnetwork from then on.

Some recent works extend the LTH from random initialization to pre-trained initialization \cite{AllTicketWin,BERT-LT,SuperTickets,EarlyBERT}. Particularly, \citet{BERT-LT} find that WTs, i.e., subnetworks of the pre-trained BERT, derived from the pre-training task of MLM using IMP are universally transferable to the downstream tasks. The same question of transferring WTs found in pre-training tasks is also explored in the CV field by \citet{LTCV1,LTCV2}. EarlyBERT \cite{EarlyBERT} investigates drawing early-bird tickets of BERT. In this work, we follow the question of transferring WTs and seek to further improve the transferability of BERT subnetworks.

\subsection{BERT Compression}
In the literature of BERT compression, pruning \cite{OBD,HanNips} and KD \cite{KD} are two widely-studied techniques. BERT can be pruned in either unstructured \cite{CompressingBERT,MovementPruning,LadaBERT} or structured \cite{AttnPrun,DynaBERT} ways. Although unstructured pruning is not hardware-friendly for speedup purpose, it is a common setup in LTH, and some recent efforts have been made in sparse tensor acceleration \cite{FastSparseConvNets,EdgeBERT}. In BERT KD, various knowledge are explored, which includes the soft-labels \cite{DistillBERT}, the hidden states \cite{PKD,DynaBERT,MUD} and the attention relations \cite{TinyBERT}, among others. Usually, pruning and KD are combined to compress the fine-tuned BERT. By contrast, the LTH compresses BERT before fine-tuning.

Another way to obtain more efficient BERT with the same transferability as the original one is to pre-train a compact model from scratch. This model can be trained either with the MLM objective \cite{BERT-small} or using pre-trained BERT as the teacher to perform KD \cite{MiniLM,MobileBERT,TinyBERT}. By contrast, the LTH extracts subnetworks from BERT, which is about exposing the knowledge already learned by BERT, rather than learning new knowledge from scratch. Compared with training a new PLM, the LTH in BERT is still underexplored in the literature.

\subsection{Learning Subnetwork Structure via Binary Mask Training}
To make the subnetwork structure trainable, we need to back-propagate gradients through the binary masks. This can be achieved through the \textit{straight-through estimator} \cite{GradEstimator} or drawing the mask variables from a \textit{hard-concrete distribution} \cite{HardConcrete} and then using the re-parameterization trick. Mask training has been utilized in model compression \cite{StructuredPruneLM,MovementPruning}, and parameter-efficient training \cite{Piggyback,Masking,HowFine}. However, unlike these works that learn the mask for each task separately (\textbf{task-specific}), we learn the subnetwork structure on pre-training task and transfer it to multiple downstream tasks (\textbf{task-agnostic}).

%%%%%%%%%%%%%%%%%%%%%%%%%%%%%%%%%%%%%%%%%%%%%%
\begin{figure*}[t]
\centering
\includegraphics[width=0.8\textwidth]{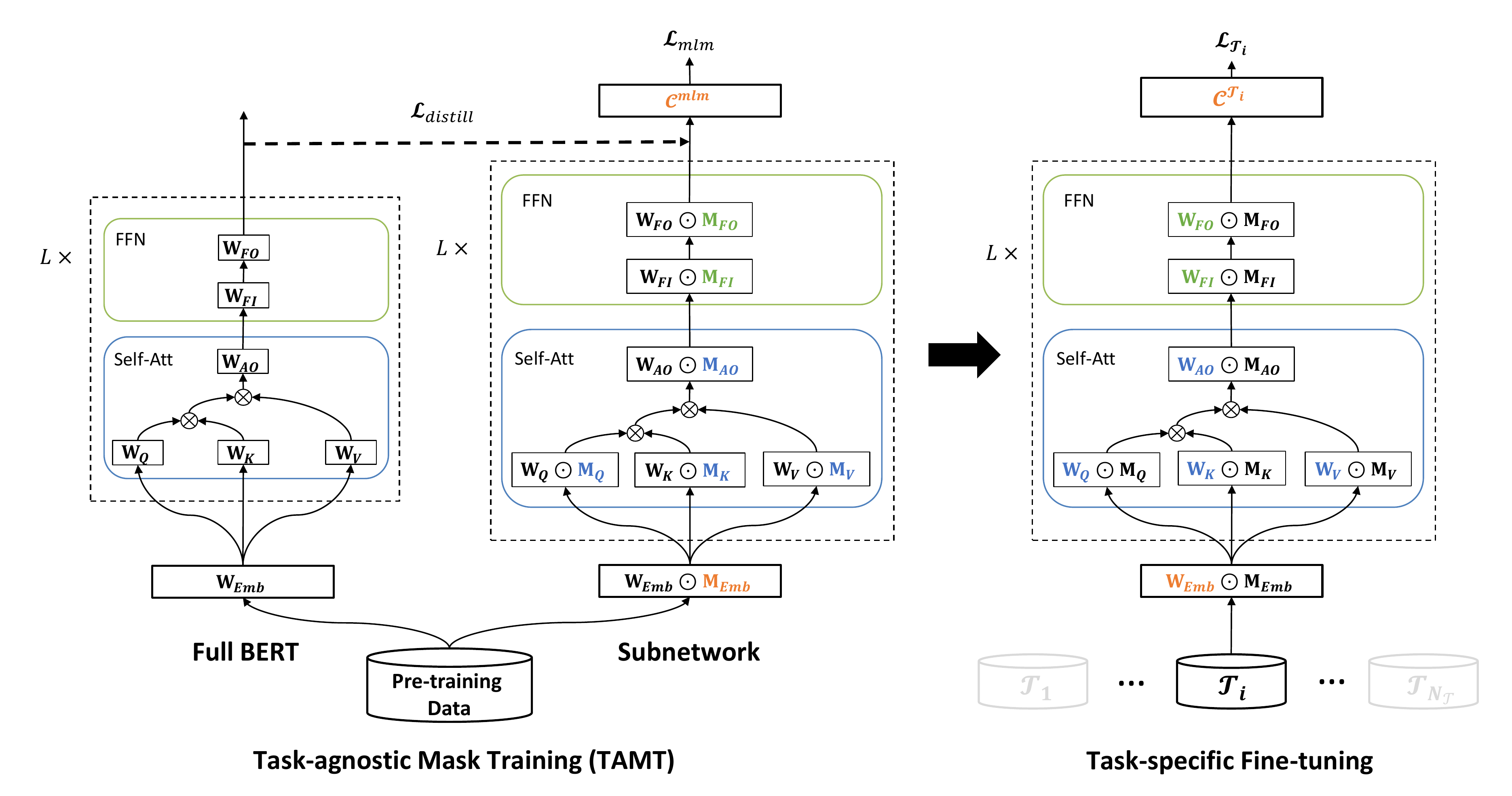}
\caption{Illustration of the BERT subnetwork transfer problem and the proposed TAMT. We search the subnetworks by training binary masks on the pre-training dataset, using either the MLM loss or the KD loss (left). The identified subnetwork is then fine-tuned on a range of downstream tasks (right). The colored weights/masks are trainable and the black ones are frozen. The residual connection and layer normalization are omitted for simplicity.}
\label{fig:method}
\end{figure*}
%%%%%%%%%%%%%%%%%%%%%%%%%%%%%%%%%%%%%%%%%%%%%%

\section{Methodology}

\subsection{BERT Architecture}
BERT consists of an embedding layer and $L$ Transformer layers \cite{Transformer}. Each Transformer layer has two sub-layers: the self-attention layer and the feed-forward network (FFN).

The self-attention layer contains $N_{h}$ parallel attention heads and each head can be formulated as:
\begin{equation}
\resizebox{1.0\hsize}{!}{$
    \operatorname{Self-Att}_{h}(\mathbf{H})=\operatorname{softmax}\left(\frac{(\mathbf{H} \mathbf{W}_{Q_h}) (\mathbf{H} \mathbf{W}_{K_h})^{\top}}{\sqrt{d_{h}}}\right) \mathbf{H} \mathbf{W}_{V_h}
$}
\end{equation}
where $\mathbf{H} \in \mathbb{R}^{|\mathbf{x}| \times d_{H}}$ is the input; $d_{H}$ and $|\mathbf{x}|$ are the hidden size and the length of input $\mathbf{x}$, respectively. $\mathbf{W}_{Q_h,K_h,V_h} \in \mathbb{R}^{d_{H} \times d_{h}}$ are the query, key and value matrices, and $d_{h}=\frac{d_{H}}{N_{h}}$. In practice, the matrices for different heads will be combined into three large matrices $\mathbf{W}_{Q,K,V} \in \mathbb{R}^{d_{H} \times d_{H}}$. The outputs of the $N_{h}$ heads are then concatenated and linearly projected by $\mathbf{W}_{AO} \in \mathbb{R}^{d_{H} \times d_{H}}$ to obtain the final output of the self-attention layer.

The FFN consists of two weight matrices $\mathbf{W}_{FI} \in \mathbb{R}^{d_{H} \times d_{I}}$, $\mathbf{W}_{FO} \in \mathbb{R}^{d_{I} \times d_{H}}$ with a GELU activation \cite{gelu} in between, where $d_I$ is the hidden dimension of FFN. Dropout \cite{Dropout}, residual connection \cite{ResNet} and layer normalization \cite{layer-norm} are also applied following each sub-layer. Eventually, for each downstream task, a classifier is used to give the final prediction based on the output of the Transformer module.

\subsection{Subnetwork and Magnitude Pruning}
Consider a model $f(\cdot ; \boldsymbol{\boldsymbol{\theta}})$ with weights $\boldsymbol{\theta}$, we can obtain its subnetwork $f(\cdot ; \mathbf{M} \odot \boldsymbol{\theta})$ by applying a binary mask $\mathbf{M} \in\{0,1\}^{|\boldsymbol{\theta}|}$ to $\boldsymbol{\theta}$, where $\odot$ denotes element-wise multiplication. In terms of BERT, we extract the subnetwork from the pre-trained weights $\boldsymbol{\theta}_0$. Specifically, we consider the matrices of the Transformer sub-layers and the word embedding matrix, i.e., $\boldsymbol{\theta}_{0} = \left\{\mathbf{W}_{E m b}\right\} \cup\left\{\mathbf{W}_{Q}^{l}, \mathbf{W}_{K}^{l}, \mathbf{W}_{V}^{l}, \mathbf{W}_{A O}^{l}, \mathbf{W}_{F I}^{l}, \mathbf{W}_{F O}^{l}\right\}_{l=1}^{L}$.

Magnitude pruning \cite{HanNips} is initially used to compress a trained neural network by setting the low-magnitude weights to zero. It can be conducted in two different ways: 1) \textit{Oneshot magnitude pruning} (OMP) directly prunes the trained weights to target sparsity while 2) \textit{iterative magnitude pruning} (IMP) performs pruning and re-training iteratively until reaching the target sparsity. OMP and IMP are also widely studied in the literature of LTH as the method to find the matching subnetworks, with an additional operation of resetting the weights to initialization.

\subsection{Problem Formulation: Transfer BERT Subnetwork}
As depicted in Fig. \ref{fig:method}, given $N_{\mathcal{T}}$ downstream tasks $\mathcal{T} = \left\{\mathcal{T}_{i}\right\}_{i=1}^{N_{\mathcal{T}}}$, the subnetwork $f\left(\cdot ; \mathbf{M} \odot \boldsymbol{\theta}_{0}, \mathcal{C}^{\mathcal{T}_{i}}_{0}\right)$ is fine-tuned on each task, together with the randomly initialized task-specific linear classifier $\mathcal{C}^{\mathcal{T}_{i}}_{0}$. We formulate the training algorithm for task $\mathcal{T}_{i}$ as a function $\mathcal{A}_{t}^{\mathcal{T}_i}\left(f\left(\cdot ; \mathbf{M} \odot \boldsymbol{\theta}_{0}, \mathcal{C}^{\mathcal{T}_{i}}_{0}\right)\right)$ (e.g., Adam or SGD), which trains the model for $t$ steps and produces $f\left(\cdot ; \mathbf{M} \odot \boldsymbol{\theta}_{t}, \mathcal{C}^{\mathcal{T}_{i}}_{t}\right)$. After fine-tuning, the model is evaluated against the metric $\mathcal{E}^{\mathcal{T}_i}(f(\cdot ; \mathbf{M} \odot \boldsymbol{\theta}_t, \mathcal{C}^{\mathcal{T}_{i}}_{t}))$ (e.g., Accuracy or F1) for task $\mathcal{T}_{i}$.

In this work, we focus on finding a BERT subnetwork, that maximally preserves the overall downstream performance given a particular sparsity $\mathcal{S}$, especially at the sparsity that magnitude pruning performs poorly. This can be formalized as:
\begin{equation}
\begin{gathered}
\max _{\mathbf{M}}\left(\frac{1}{N_{\mathcal{T}}}\sum_{i=1}^{N_{\mathcal{T}}} \mathcal{E}^{\mathcal{T}_i}\left(\mathcal{A}_{t}^{\mathcal{T}_i}\left(f \left(\cdot, \mathbf{M} \cdot \boldsymbol{\theta}_{0}, \mathcal{C}_{0}^{\mathcal{T}_i}\right)\right)\right)\right)  \\
\text { s.t. } \frac{\|\mathbf{M}\|_{0}}{|\boldsymbol{\theta}_0|} =(1-\mathcal{S})
\end{gathered}
\end{equation}
where $\|\mathbf{M}\|_{0}$ and $|\boldsymbol{\theta}_0|$ are the $L_0$ norm of the mask and the total number of model weights respectively.

\subsection{Task-agnostic Mask Training}
\subsubsection{Mask Training with Binarization and Gradient Estimation}
In order to learn the binary masks, we adopt the technique for training binarized neural networks \cite{BinaryNet}, following \citet{Masking,Piggyback}. This technique involves mask binarization in the forward pass and gradient estimation in the backward pass.

As shown in Fig. \ref{fig:method}, each weight matrix $\mathbf{W} \in \mathbb{R}^{d_{in} \times d_{out}}$ is associated with a binary mask $\mathbf{M} \in\{0,1\}^{d_{in} \times d_{out}}$, which is derived from a real-valued matrix $\overline{\mathbf{M}} \in \mathbb{R}^{d_{in} \times d_{out}}$ via binarization:
\begin{equation}
\label{eq:estimation}
\mathbf{M}_{i, j}= \begin{cases}1 & \text { if } \overline{\mathbf{M}}_{i, j} \geq \phi \\ 0 & \text { otherwise }\end{cases}
\end{equation}
where $\phi$ is the threshold that controls the sparsity. In the forward pass of a subnetwork, $\mathbf{W} \odot \mathbf{M}$ is used in replacement of the original weights $\mathbf{W}$.

Since $\mathbf{M}_{i, j}$ are discrete variables, the gradient signals cannot be back-propagated through the binary mask. We therefore use the \textit{straight-through estimator} \cite{GradEstimator} to approximate the gradients and update the real-valued mask:
\begin{equation}
\overline{\mathbf{M}} \leftarrow \overline{\mathbf{M}}-\eta \frac{\partial \mathcal{L}}{\partial \mathbf{M}}
\end{equation}
where $\mathcal{L}$ is the loss function and $\eta$ is the learning rate. In other words, the gradients of $\overline{\mathbf{M}}$ is estimated using the gradients of $\mathbf{M}$. In the process of mask training, all the original weights are frozen.

\subsubsection{Mask Initialization and Sparsity Control}
\label{sec:mask_init}
The real-valued masks can be initialized in various forms, e.g., random initialization. Considering that magnitude pruning can preserve the pre-training knowledge to some extent, and OMP is easy to implement with almost zero computation cost, we directly initialize $\overline{\mathbf{M}}$ using OMP:
\begin{equation}
\label{eq:omp_init}
\overline{\mathbf{M}}_{i, j}= \begin{cases} \alpha \times \phi & \text { if } \mathbf{M}^{OMP}_{i, j} = 1 \\ 0 & \text { otherwise }\end{cases}
\end{equation}
where $\mathbf{M}^{OMP}$ is the binary mask derived from OMP and $\alpha \geq 1$ is a hyper-parameter. In this way, the weights with large magnitudes will be retained at initialization according to Eq. \ref{eq:estimation}, because the corresponding $\overline{\mathbf{M}}_{i, j} = \alpha \times \phi \geq \phi$. In practice, we perform OMP over the weights \textit{locally} based on the given sparsity, which means the magnitudes are ranked inside each weight matrix. 

As $\overline{\mathbf{M}}$ being updated, some of its entries with zero initialization will gradually surpass the threshold, and vice versa. If the threshold $\phi$ is fixed throughout training, there is no guarantee that the binary mask will always satisfy the given sparsity. Therefore, we rank $\overline{\mathbf{M}}_{i j}$ according to their absolute values during mask training, and dynamically adjust the threshold to satisfy the sparsity constraint.

\subsubsection{Mask Training Objectives}
We explore the use of two objectives for mask training, namely the MLM loss and the KD loss.

The MLM is the original task used in BERT pre-training. It randomly replaces a portion of the input tokens with the $[\mathrm{MASK}]$ token, and requires the model to reconstruct the original tokens based on the entire masked sequence. Concretely, the MLM objective is computed as cross-entropy loss on the predicted masked tokens. During MLM learning, we allow the token classifier (i.e., the $\mathcal{C}^{mlm}$ in Fig. \ref{fig:method}) to be trainable, in addition to the masks.

In KD, the compressed model (student) is trained with supervision from the original model (teacher). Under our framework of mask training, the training signal can also be derived from the unpruned BERT. To this end, we design the KD objective by encouraging the subnetwork to mimic the representations of the original BERT, which is shown to be a useful source of knowledge in BERT KD \cite{PKD,DynaBERT}. Specifically, the distillation loss is formulated as the cosine similarity between the teacher's and student's representations:
\begin{equation}
\mathcal{L}_{distill}=\frac{1}{L|\mathbf{x}|} \sum_{l=1}^{L} \sum_{i=1}^{|\mathbf{x}|} (1-\cos \left(\mathbf{H}_{l, i}^{T}, \mathbf{H}_{l, i}^{S}\right) )
\label{eq:loss_distill}
\end{equation}
where $\mathbf{H}_{l, i}$ is the hidden state of the $i^{th}$ token at the $l^{th}$ layer; $T$ and $S$ denote the teacher and student respectively; $\cos(\cdot,\cdot)$ is the cosine similarity.

\section{Experiments}

\subsection{Experimental Setups}
\subsubsection{Models}
We examine two PLMs from the BERT family, i.e., $\mathrm{BERT}_{\mathrm{BASE}}$ \cite{BERT} and $\mathrm{RoBERTa}_{\mathrm{BASE}}$ \cite{RoBERTa}. They have basically the same structure, while differ in the vocabulary size, which results in approximately 110M and 125M parameters respectively. The main results of Section \ref{sec:main_result} study both two models. For the analytical studies, we only use $\mathrm{BERT}_{\mathrm{BASE}}$.

%%%%%%%%%%%%%%%%%%%%%%%%%%%%%%%%%%%%%%%%%%%%%%
\begin{figure*}[t]
\centering
\includegraphics[width=0.9\textwidth]{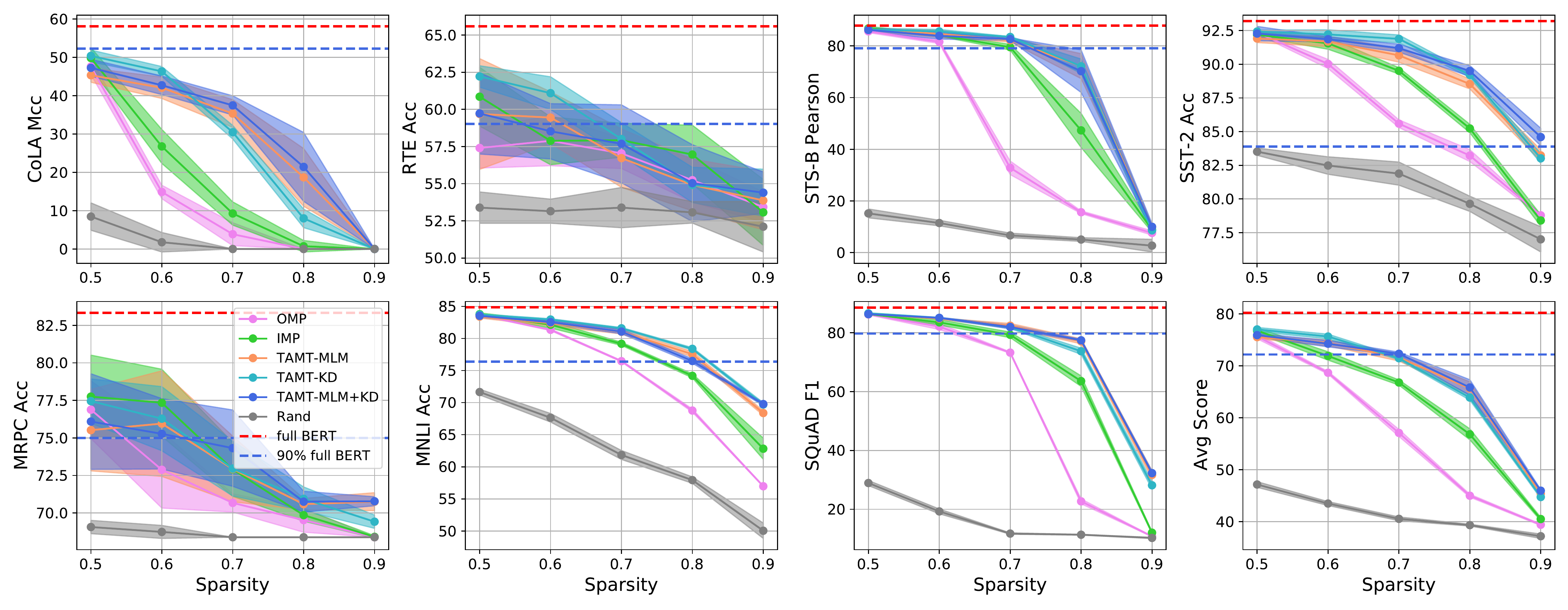}
\caption{Downstream performance of BERT subnetworks. Shadowed areas denote standard deviations.}
\label{fig:main_result_bert}
\end{figure*}
%%%%%%%%%%%%%%%%%%%%%%%%%%%%%%%%%%%%%%%%%%%%%%
%%%%%%%%%%%%%%%%%%%%%%%%%%%%%%%%%%%%%%%%%%%%%%
\begin{figure*}[t]
\centering
\includegraphics[width=0.9\textwidth]{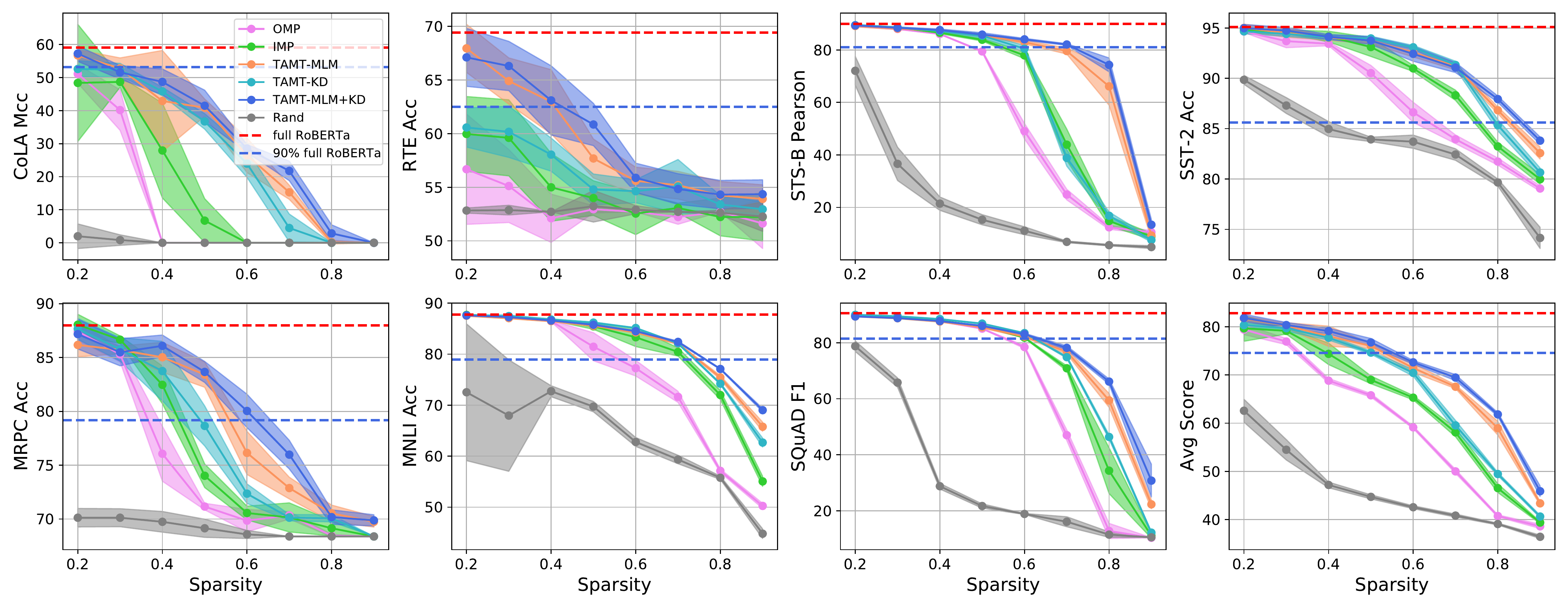}
\caption{Downstream performance of RoBERTa subnetworks. Shadowed areas denote standard deviations.}
\label{fig:main_result_roberta}
\end{figure*}
%%%%%%%%%%%%%%%%%%%%%%%%%%%%%%%%%%%%%%%%%%%%%%
\subsubsection{Baselines, Datasets and Evaluation}
We compare our mask training method with IMP, OMP as well as subnetworks with random structures. Following \citet{BERT-LT}, we use the MLM loss during IMP training. For TAMT, we consider three variants, namely TAMT-MLM that uses MLM as training objective, TAMT-KD that uses the KD objective (Eq. \ref{eq:loss_distill}), and TAMT-MLM+KD that equally combines MLM and KD.
% As for OMP and the random subnetworks, they do not require training.

We build our pre-training set using the WikiText-103 dataset \cite{WikiText} for language modeling. For downstream fine-tuning, we use six datasets, i.e., CoLA, SST-2, RTE, MNLI, MRPC and STS-B from the GLUE benchmark for NLU and the SQuAD v1.1 dataset for QA.

Evaluations are conducted on the dev sets. For the downstream tasks, we follow the standard evaluation metrics \cite{GLUE}. For the pre-training tasks, we calculate the MLM and KD loss on the dev set of WikiText-103. More information about the datasets and evaluation metrics can be found in Appendix \ref{sec:appendix_b1}.

\subsubsection{Implementation Details}
Both TAMT and IMP are conducted on the pre-training dataset. For mask training, we initialize the mask using OMP as described in Section \ref{sec:mask_init}. We also provide a comparison between OMP and random initialization in Section \ref{sec:mask_init_experiment}. The initial threshold $\phi$ and $\alpha$ are set to $0.01$ and 2 respectively, which work well in our experiments. For IMP, we increase the sparsity by $10\%$ every 1/10 of total training iterations, until reaching the target sparsity, following \citet{BERT-LT}. Every pruning operation in IMP is followed by resetting the remaining weights to $\theta_0$. In the fine-tuning stage, all the models are trained using the same set of hyper-parameters unless otherwise specified.

For TAMT, IMP and random pruning, we generate three subnetworks with different seeds, and the result of each subnetwork is also averaged across three runs, i.e., the result of every method is the average of nine runs in total. For OMP, we can only generate one subnetwork, which is fine-tuned across three runs. More implementation details and computing budgets can be found in Appendix \ref{sec:appendix_b2}.

%%%%%%%%%%%%%%%%%%%%%%%%%%%%%%%%%%%%%%%%%%%%%%
\begin{figure*}[t]
\centering
\includegraphics[width=0.85\textwidth]{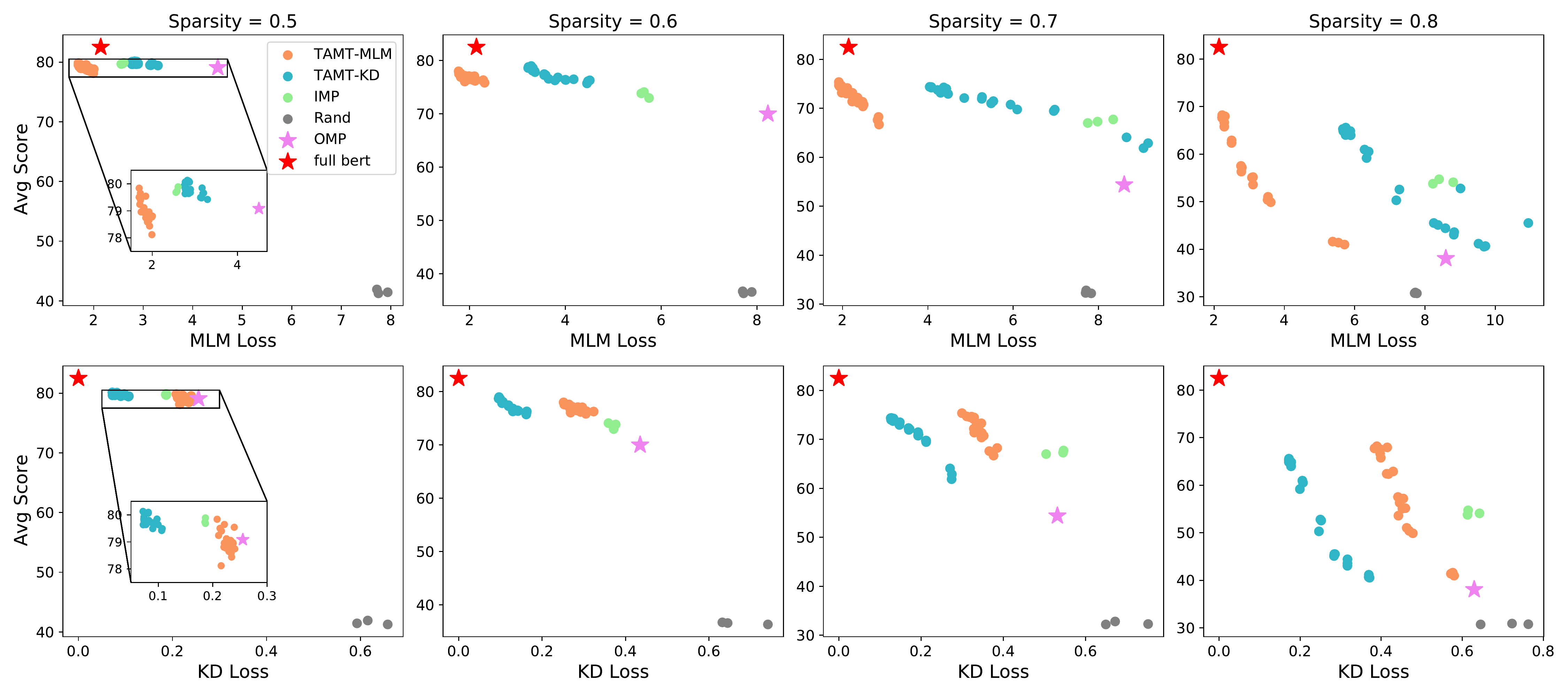}
\caption{MLM/KD dev loss and downstream results. The results of TAMT are from the masks along the training process, and the results of IMP and Rand are from different seeds. Appendix \ref{sec:appendix_a3} shows the results on each task.}
\label{fig:mask_loss}
\end{figure*}
%%%%%%%%%%%%%%%%%%%%%%%%%%%%%%%%%%%%%%%%%%%%%%

\subsection{Results and Analysis}
\subsubsection{Main Results}
\label{sec:main_result}
Fig. \ref{fig:main_result_bert} and Fig. \ref{fig:main_result_roberta} present the downstream performance of BERT and RoBERTa subnetworks, respectively. We can derive the following observations:

There is a clear gap between random subnetworks and the other ones found with certain inductive bias. At $50\%$ sparsity for BERT and $30\%$ for RoBERTa, all the methods, except for ``Rand'', maintain $90\%$ of the full model's overall performance. As sparsity grows, the OMP subnetworks degrade significantly. IMP, which is also based on magnitude, exhibits relatively mild declines.

TAMT further outperforms IMP with perceivable margin. For BERT subnetworks, the performance of TAMT variants are close to each other, which have advantage over IMP across $60\%\sim90\%$ sparsity. When it comes to RoBERTa, the performance of TAMT-KD is undesirable at $70\%\sim90\%$ sparsity, which only slightly outperforms IMP. In comparison, TAMT-MLM consistently surpasses IMP and TAMT-KD on RoBERTa.

Combining MLM and KD leads to comparable average performance as TAMT-MLM for BERT, while slightly improves over TAMT-MLM for RoBERTa. This suggests that the two training objectives could potentially benefit, or at least will not negatively impact each other. In Section \ref{sec:pretrain_performance_transferability}, we will show that the MLM and KD objectives indeed exhibit certain consistency.

At $90\%$ sparsity, all the methods perform poorly, with average scores approximately half of the full model. On certain tasks like CoLA, RTE and MRPC, drastic performance drop of all methods can even be observed at lower sparsity (e.g., $60\%\sim80\%$). This is probably because the number of training data is too scarce in these tasks for sparse PLMs to perform well. However, we find that the advantage of TAMT is more significant within a range of data scarsity, which will be discussed in Section \ref{sec:few-shot}.

We also note that RoBERTa, although outperforms BERT as a full model, is more sensitive to task-agnostic pruning. A direct comparison between the two PLMs is provided in Appendix \ref{sec:appendix_a2}.

\subsubsection{The Effect of Pre-training Performance}
\label{sec:pretrain_performance_transferability}
As we discussed in Section \ref{sec:intro}, our motivation of mask training is to improve downstream transferability by preserving the pre-training performance. To examine whether the effectiveness of TAMT is indeed derived from the improvement on pre-training tasks, we calculate the MLM/KD dev loss for the subnetworks obtained from the mask training process, and associate it with the downstream performance. The results are shown in Fig. \ref{fig:mask_loss}, where the "Avg Score" includes CoLA, SST-2, MNLI, STS-B and SQuAD. In the following sections, we also mainly focus on these five tasks. We can see from Fig. \ref{fig:mask_loss} that:

There is a positive correlation between the pre-training and downstream performance, and this trend can be observed for subnetworks across different sparsities. Compared with random pruning, the magnitude pruning subnetworks and TAMT subnetworks reside in an area with lower MLM/KD loss and higher downstream score at $50\%$ sparsity. As sparsity increases, OMP subnetworks gradually move from the upper-left to the lower-right area of the plots. In comparison, IMP is better at preserving the pre-training performance, even though it is not deliberately designed for this purpose. For this reason, hypothetically, the downstream performance of IMP is also better than OMP. 

TAMT-MLM and TAMT-KD have the lowest MLM and KD loss respectively, which demonstrates that the masks are successfully optimized towards the given objectives. As a result, the downstream performance is also elevated from the OMP initialization, which justifies our motivation. Moreover, training the mask with KD loss can also optimize the performance on MLM, and vice versa, suggesting that there exists some consistency between the objectives of MLM and KD.

It is also worth noting that the correlation between pre-training and fine-tuning performance is not ubiquitous. For example, among the subnetworks of OMP, IMP and TAMT at $50\%$ sparsity, the decrease in KD/MLM loss produces little or no downstream improvement; at $60\% \sim 80\%$ sparsity, OMP underperforms random pruning in MLM, while its downstream performance is better. These phenomenons suggest that some properties about the BERT winning tickets are still not well-understood by us.

%%%%%%%%%%%%%%%%%%%%%%%%%%%%%%%%%%%%%%%%%%%%%%
\begin{figure}[t]
\centering
\includegraphics[width=1.0\linewidth]{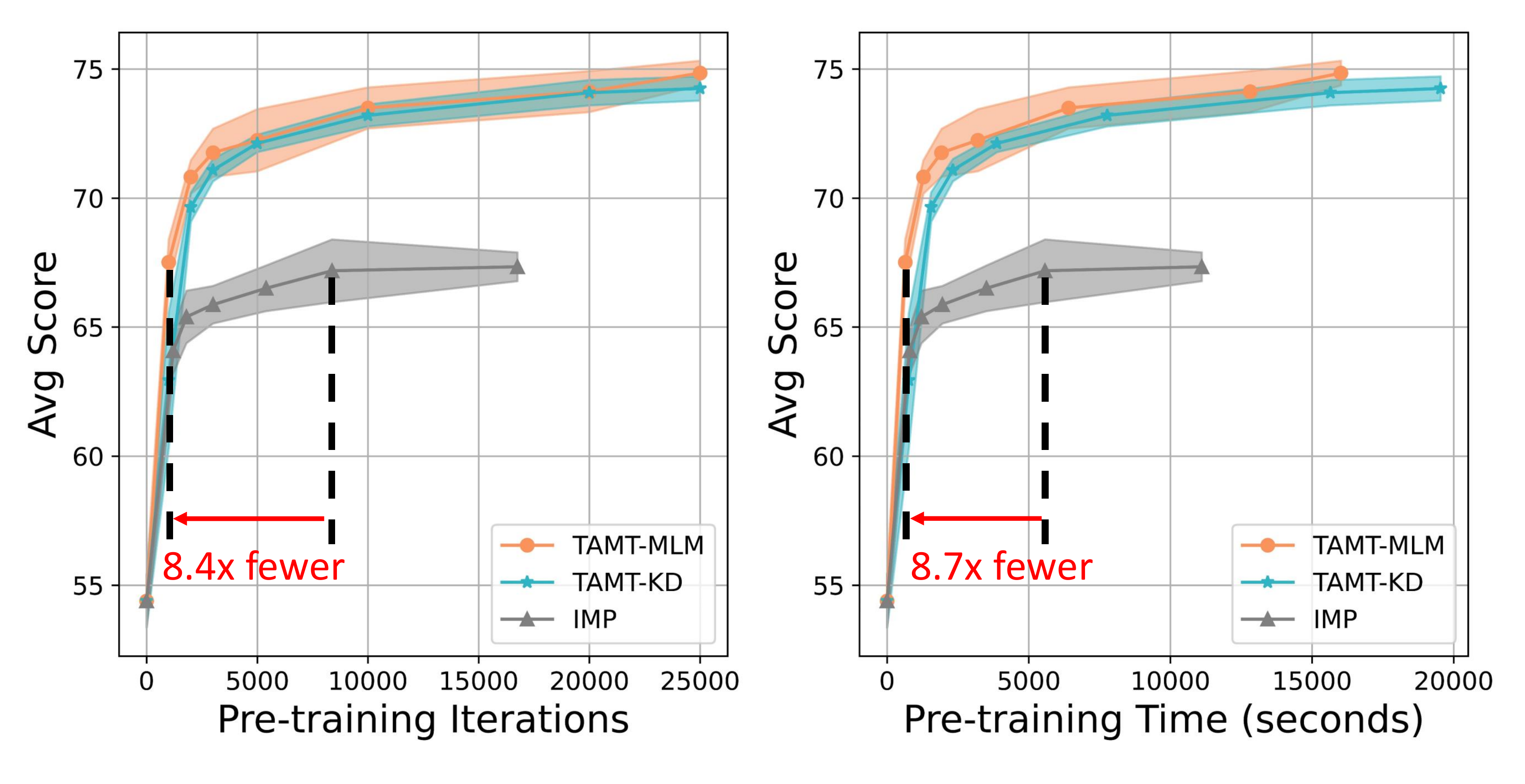}
\caption{The downstream performance of masks at $70\%$ sparsity with increased pre-training cost. The training time is computed excluding evaluation. Shadowed areas denote standard deviations. Results for each task and more sparsities are shown in Appendix \ref{sec:appendix_a4}.}
\label{fig:mask_step}
\end{figure}
%%%%%%%%%%%%%%%%%%%%%%%%%%%%%%%%%%%%%%%%%%%%%%
%%%%%%%%%%%%%%%%%%%%%%%%%%%%%%%%%%%%%%%%%%%%%%
\begin{figure}[t]
\centering
\includegraphics[width=1.\linewidth]{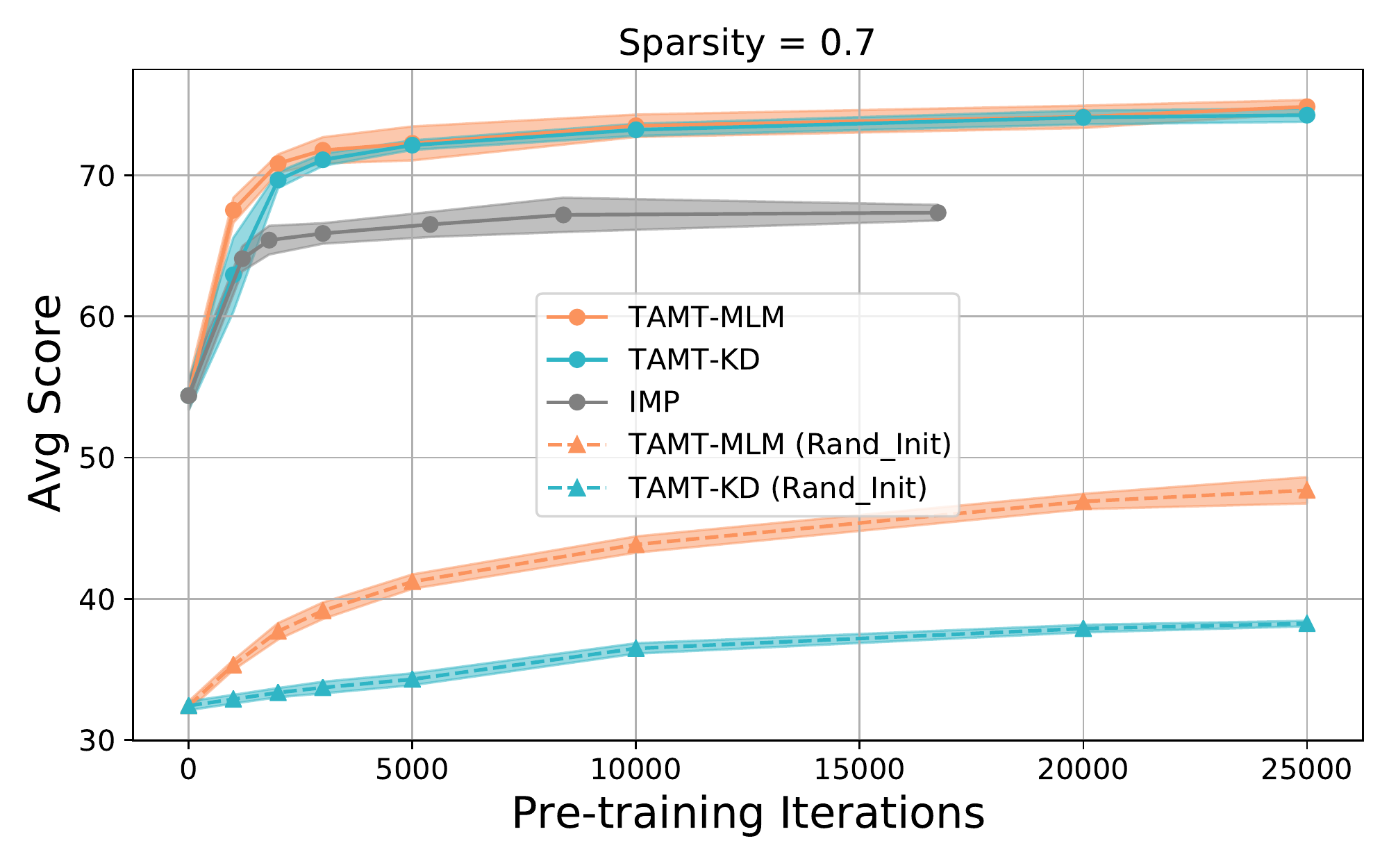}
\caption{Comparison between OMP initialization (solid lines) and random initialization (dashed lines) of masks at $70\%$ sparsity. The axes are defined in the same way as the left plot of Fig. \ref{fig:mask_step}.}
\label{fig:mask_init}
\end{figure}
%%%%%%%%%%%%%%%%%%%%%%%%%%%%%%%%%%%%%%%%%%%%%%
%%%%%%%%%%%%%%%%%%%%%%%%%%%%%%%%%%%%%%%%%%%%%%
\begin{figure}[t]
\centering
\includegraphics[width=1.\linewidth]{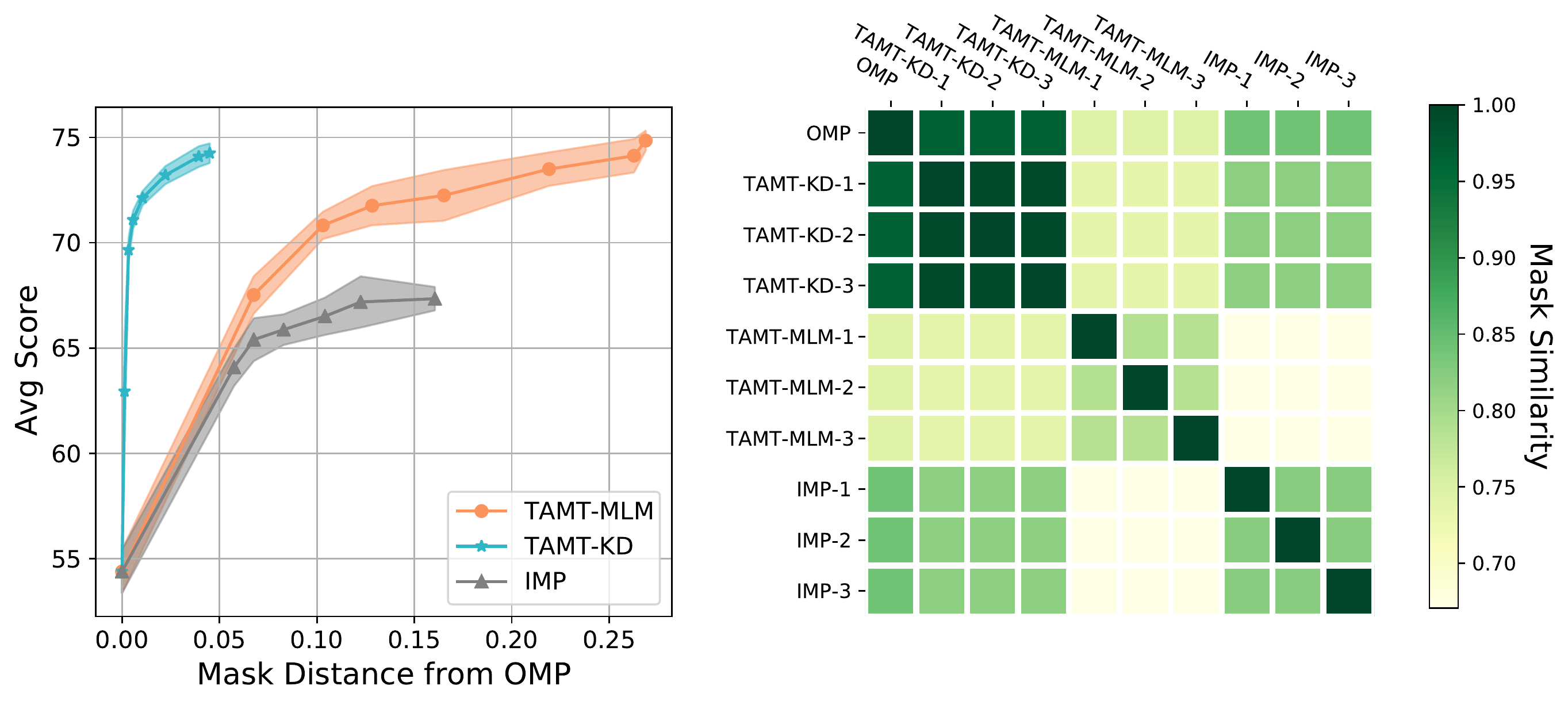}
\caption{Left: The downstream results of masks with varying distances from the OMP mask. Shadowed areas denote standard deviations. Right: The similarity between the masks used to report the main results at $70\%$ sparsity. The suffix numbers indicate different seeds. Results of more sparsities are shown in Appendix \ref{sec:appendix_a5}.}
\label{fig:mask_sim}
\end{figure}
%%%%%%%%%%%%%%%%%%%%%%%%%%%%%%%%%%%%%%%%%%%%%%

\subsubsection{The Effect of Pre-training Cost}
We have shown that mask training is more effective than magnitude pruning. Now let us take a closer look at the results of TAMT and IMP with different iterations of pre-training, to evaluate their efficiency in subnetwork searching. For TAMT, we directly obtain the subnetworks from varied pre-training iterations. For IMP, we change the pruning frequency to control the number of training iterations before reaching the target sparsity.

Fig. \ref{fig:mask_step} presents the downstream results with increased pre-training iterations and time. We can see that for all the methods, the fine-tuning performance steadily improves as pre-training proceeds. Along this process, TAMT advances at a faster pace, reaching the best score achieved by IMP with $8.4\times$ fewer iterations and $8.7\times$ fewer time. This indicates that directly optimizing the pre-training objectives is more efficient than the iterative process of weight pruning and re-training.

\subsubsection{The Effect of Mask Initialization}
\label{sec:mask_init_experiment}
In the main results, we use OMP as the default initialization, in order to provide a better start point for TAMT. To validate the efficacy of this setting, we compare OMP initialization with random initialization. Concretely, we randomly sample some entries of the real-valued masks to be zero, according to the given sparsity, and use the same $\alpha$ and $\phi$ for the non-zero entries as in Eq. \ref{eq:omp_init}.

The results are shown in Fig. \ref{fig:mask_init}. We can see that, for random initialization, TAMT can still steadily improve the downstream performance as pre-training proceeds. However, the final results of TAMT-MLM/KD (Rand\_init) are significantly worse than TAMT-MLM/KD, which demonstrates the necessity of using OMP as initialization.

\subsubsection{Similarity between Subnetworks}
\label{sec:mask_dist_sim}
The above results show that the subnetworks found by different methods perform differently. We are therefore interested to see how they differ in the mask structure. To this end, we compute the similarity between OMP mask and the masks derived during the training of TAMT and IMP. Following \citet{BERT-LT}, we measure the Jaccard similarity between two binary masks $\mathbf{M}_i$ and $\mathbf{M}_j$ as $\frac{|\mathbf{M}_{i} \cap \mathbf{M}_{j}|}{|\mathbf{M}_{i} \cup \mathbf{M}_{j}|}$, and the \textit{mask distance} is defined as $1 - \frac{|\mathbf{M}_{i} \cap \mathbf{M}_{j}|}{|\mathbf{M}_{i} \cup \mathbf{M}_{j}|}$.

From the results of Fig. \ref{fig:mask_sim}, we can find that: 1) With different objectives, TAMT produces different mask structures. The KD loss results in masks in the close proximity of OMP initialization, while the MLM masks deviate away from OMP. 2) Among the four methods, IMP and TAMT-MLM have the highest degree of dissimilarity, despite the fact that they both involve MLM training. 3) Although IMP, TAMT-KD and TAMT-MLM are different from each other in terms of subnetwork structure, all of them clearly improves over the OMP baseline. Therefore, we hyphothesize that the high-dimensional binary space $\{0, 1\}^{|\boldsymbol{\theta}|}$ might contain multiple regions of winning tickets that are disjoint with each other. Searching methods with different inductive biases (e.g., mask training versus pruning and KD loss versus MLM loss) are inclined to find different regions of interest.

%%%%%%%%%%%%%%%%%%%%%%%%%%%%%%%%%%%%%%%%%%%%%%
\begin{figure}[t]
\centering
\includegraphics[width=1.\linewidth]{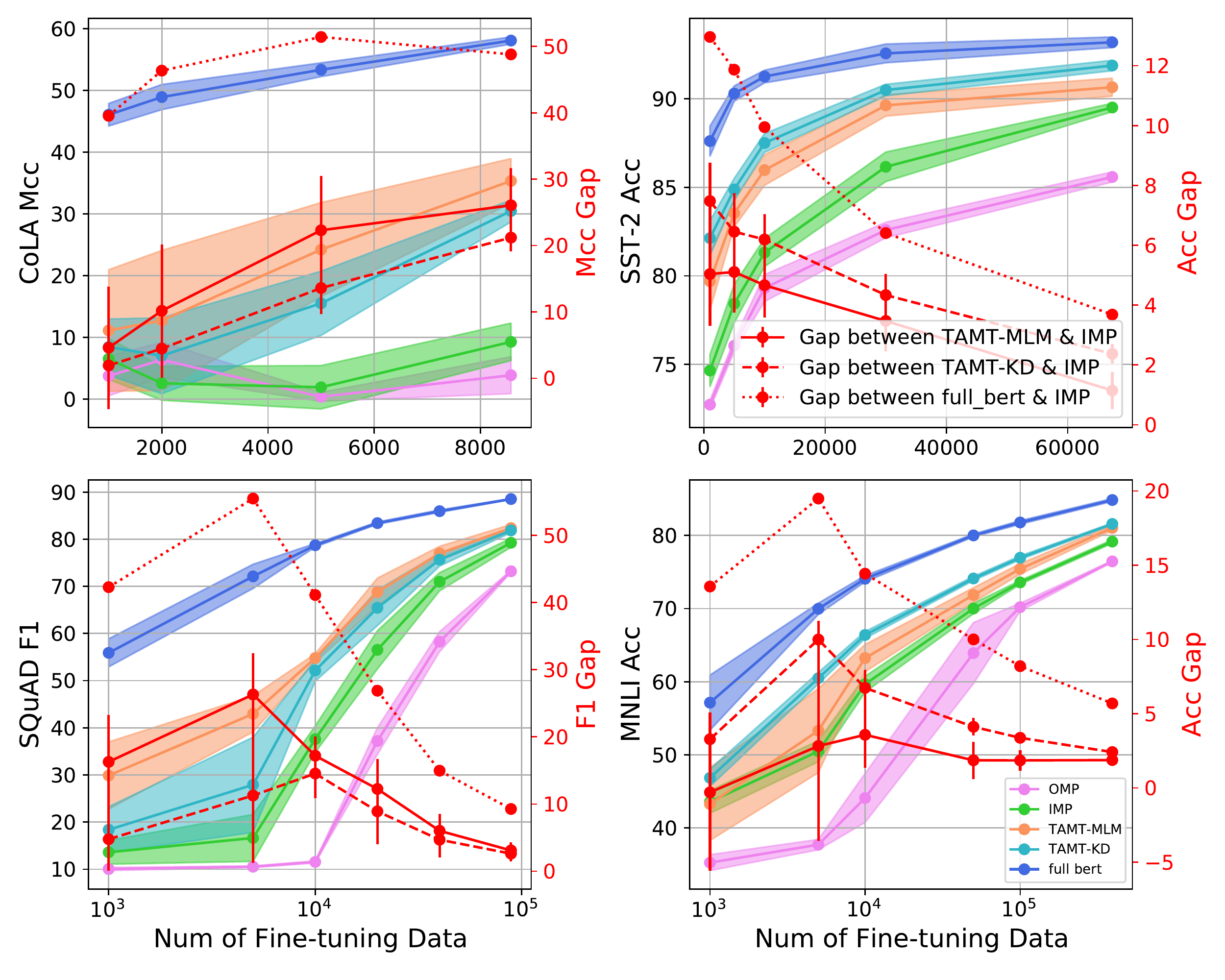}
\caption{The downstream results of full BERT and $70\%$ sparse subnetworks with varying numbers of fine-tuning data. The results are averaged over five runs for each subnetwork. Shadowed areas and error bars denote standard deviations.}
\label{fig:few_shot}
\end{figure}
%%%%%%%%%%%%%%%%%%%%%%%%%%%%%%%%%%%%%%%%%%%%%%

\subsubsection{Results of Reducing Fine-tuning Data}
\label{sec:few-shot}
To test the fine-tuning results with reduced data, we select four tasks (CoLA, SST-2, MNLI and SQuAD) with the largest data sizes and shrink them from the entire training set to 1,000 samples. 

Fig. \ref{fig:few_shot} summarizes the results of subnetworks found using different methods, as well as results of full BERT as a reference. We can see that the four datasets present different patterns. For MNLI and SQuAD, the advantage of TAMT first increases and then decreases with the reduction of data size. The turning point appears at around 10,000 samples, after which the performance of all methods, including the full BERT, degrade drastically (note that the horizontal axis is in log scale). For SST-2, the performance gap is enlarged continuously until we have only 1,000 data. With regard to CoLA, the gap between TAMT and IMP shrinks as we reduce the data size, which is not desirable. However, a decrease in the gap between full BERT and IMP is also witnessed when the data size is reduced under 5,000 samples. This is in part because the Mcc of IMP is already quite low even with the entire training set, and thus the performance decrease of IMP is limited compared with TAMT. However, the results on CoLA, as well as the results on MNLI and SQuAD with less than 10,000 samples, also suggest an inherent difficulty of learning with limited data for subnetworks at high sparsity, which is also discussed in the main results.

\section{Conclusions}
In this paper, we address the problem of searching transferable BERT subnetworks. We first show that there exist correlations between the pre-training performance and downstream transferablility of a subnetwork. Motivated by this, we devise a subnetwork searching method based on task-agnostic mask training (TAMT). We empirically show that TAMT with MLM loss or KD loss achieve better pre-training and downstream performance than the magnitude pruning, which is recently shown to be successful in finding universal BERT subnetworks. TAMT is also more efficient in mask searching and produces more robust subnetworks when being fine-tuned within a certain range of data scarsity.

\section{Limitations and Future Work}
Under the framework of TAMT, there are still some unsolved challenges and interesting questions worth studying in the future work: First, we focus on unstructured sparsity in this work, which is hardware-unfriendly for speedup purpose. In future work, we are interested in investigating TAMT with structured pruning or applying unstructured BERT subnetworks on hardware platforms that support sparse tensor acceleration \cite{FastSparseConvNets,EdgeBERT}. Second, despite the overall improvement achieved by TAMT, it fails at extreme sparsity or when the labeled data for a task is too scarce. Therefore, another future direction is to further promote the performance of universal PLM subnetworks on these challenging circumstances. To achieve this goal, thirdly, a feasible way is to explore other task-agnostic training objectives for TAMT beyond MLM and hidden state KD, e.g., self-attention KD \cite{TinyBERT} and contrastive learning \cite{SimCSE}. An in-depth study on the selection of TAMT training objective might further advance our understanding of TAMT and the LTH of BERT.

\section*{Acknowledgments}
This work was supported by the National Natural Science Foundation of China under Grants 61976207 and 61906187.

\bibliography{acl}
\bibliographystyle{acl_natbib}

\appendix
%%%%%%%%%%%%%%%%%%%%%%%%%%%%%%%%%%%%%%%%%%%%%%
\begin{figure*}[t]
\centering
\includegraphics[width=1.0\linewidth]{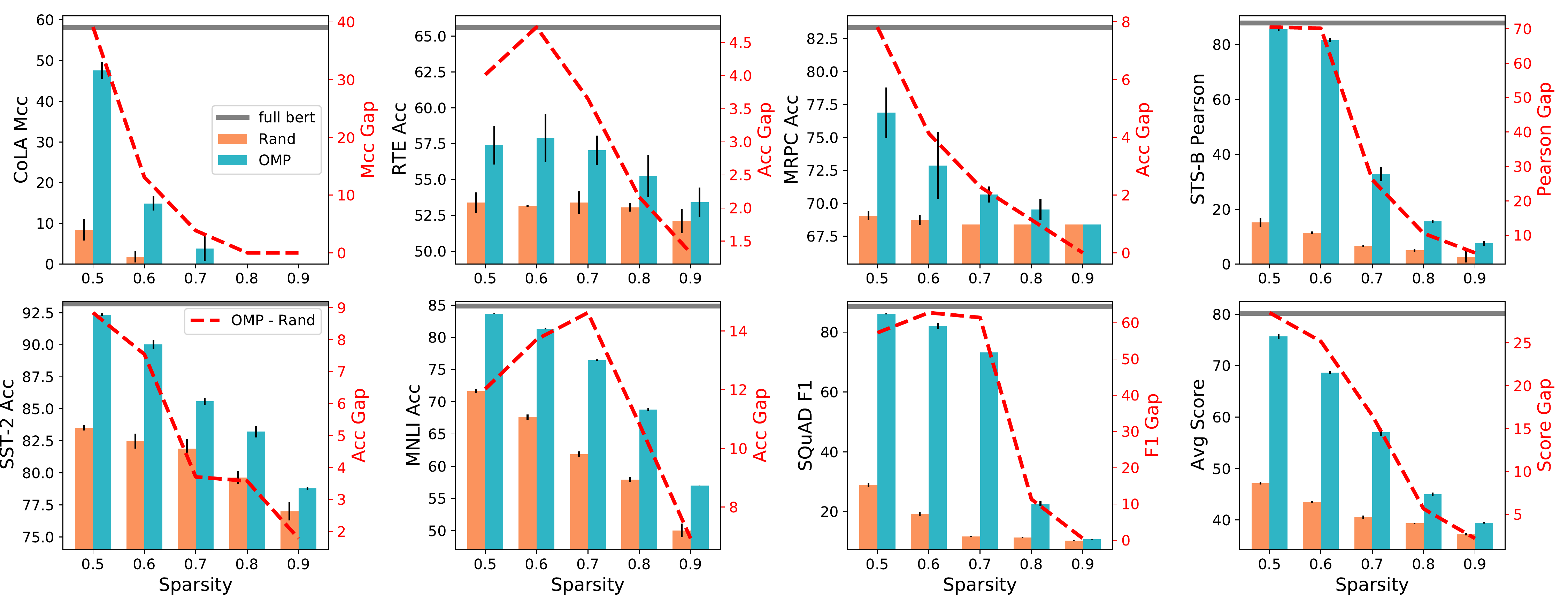}
\caption{Downstream performance of OMP subnetworks and random subnetworks of $\mathrm{BERT}_{\mathrm{BASE}}$. The error bars denote standard deviations. The dashed red line is the performance gap between ``OMP" and ``Rand".}
\label{fig:omp_rand_downstream}
\end{figure*}
%%%%%%%%%%%%%%%%%%%%%%%%%%%%%%%%%%%%%%%%%%%%%%
\section{More Results and Analysis}
\subsection{Single Task Downstream Performance of OMP and Random Pruning}
\label{sec:appendix_a1}
In Fig. \ref{fig:motivation} of the main body of paper, we show that the pre-training and overall downstream performance of OMP, as well as the gap between ``OMP" and ``Rand", degrade simultaneously as sparsity increases. The detailed results of each downstream task are presented in Fig. \ref{fig:omp_rand_downstream}. As we can see, the general pattern for every task is similar, with the exception that the gap between ``OMP" and ``Rand" slightly increases before high sparsity on tasks RTE, MNLI and SQuAD.

%%%%%%%%%%%%%%%%%%%%%%%%%%%%%%%%%%
\begin{table*}[t]
\centering
\resizebox{1\hsize}{!}{$
\begin{tabular}{@{}l c c c c c c c c c c c@{}}
\toprule
\multirow{2}{*}{} & \multicolumn{4}{c}{Pre-training} & \multicolumn{7}{c}{Fine-tuning}\\
\cmidrule(r){2-5} \cmidrule(r){6-12}
                            &IMP-MLM  &TAMT-MLM   &TAMT-KD  &TAMT-MLM+KD     &MNLI     &SST-2    &CoLA     &STS-B    &MRPC     &RTE     &SQuAD \\ \midrule
\# Train Samples            &103M     &103M       &103M     &103M            &392K     &67K      &8.5K     &5.7K     &3.6K     &2.4K    &88K   \\
\# Eval Samples             &217K     &217K       &217K     &217K            &9.8K     &0.8K     &1K       &1.5K     &0.4K     &0.2K    &10K   \\
Max Epochs                  &2        &-          &-        &-               &3        &3        &3        &3        &3        &3       &2     \\
Eval Iter                   &-        &-          &-        &-               &500      &50       &50       &50       &50       &50      &1K     \\
Batch Size                  &16       &16         &16       &16              &32       &32       &32       &32       &32       &32      &16     \\
Max Length                  &512      &512        &512      &512             &128      &128      &128      &128      &128      &128     &384     \\
Lr (linear decay)           &5e-5     &5e-5       &2e-5     &5e-5            &2e-5     &2e-5     &2e-5     &2e-5     &2e-5     &2e-5    &3e-5     \\
Eval Metric                 &Dev Loss  &Dev Loss  &Dev Loss &-               &Matched Acc   &Acc   &Matthew’s Corr   &Pearson Corr   &Acc   &Acc  &F1     \\
Optimizer                                                 & \multicolumn{11}{c}{AdamW \cite{AdamW}}  \\
\bottomrule
\end{tabular}
$}
\caption{Experimental details about IMP, task-agnostic mask training (TAMT) and fine-tuning. For pre-training, we report the number of tokens as ``\# of Train/Eval Samples''. ``Dev Loss'' denotes the loss of MLM or KD on the dev set. During fine-tuning, evaluation is performed every ``Eval Iter'' training iterations.}
\label{tab:setup}
\end{table*}
%%%%%%%%%%%%%%%%%%%%%%%%%%%%%%%%%%%%%%%%%%%
%%%%%%%%%%%%%%%%%%%%%%%%%%%%%%%%%%
\begin{table*}[t]
\centering
\resizebox{0.7 \hsize}{!}{$
\begin{tabular}{@{}l l l l l l l l l@{}}
% \begin{tabular}{@{}l c c c c c c c c c c@{}}
\toprule           &20\%     &30\%       &40\%       &50\%     &60\%       &70\%        &80\%     &90\%          \\ \midrule
IMP                &2.79K    &5.58K      &8.38K      &11.17K   &13.96K     &16.75K      &19.54K   &22.34K         \\
TAMT               &3K       &6K         &8K         &11K      &14K        &17K         &20K      &22K             \\
\bottomrule
\end{tabular}
$}
\caption{Pre-training iterations for IMP and TAMT subnetworks at $20\% \sim 90\%$ sparsity.}
\label{tab:pretrain_iterations}
\end{table*}
%%%%%%%%%%%%%%%%%%%%%%%%%%%%%%%%%%%%%%%%%%%
%%%%%%%%%%%%%%%%%%%%%%%%%%%%%%%%%%
\begin{table*}[t]
\centering
\resizebox{0.5 \hsize}{!}{$
\begin{tabular}{@{}l l l l@{}}
% \begin{tabular}{@{}l c c c c c c c c c c@{}}
\toprule                                      &IMP           &TAMT-MLM      &TAMT-KD             \\ \midrule
$\mathrm{BERT}_{\mathrm{BASE}}$                &4h6m26s       &3h54m58s      &4h46m46s               \\
$\mathrm{RoBERTa}_{\mathrm{BASE}}$             &4h33m9s       &4h17m15s      &4h51m55s                 \\
\bottomrule
\end{tabular}
$}
\caption{Pre-training time (w/o evaluation during training) of IMP and TAMT on a single on a single 32GB Nvidia V100 GPU. ``h'', ``m'' and ``s'' denote hour, minute and second, respectively. The pre-training iterations are 22.34K and 22K for IMP and TAMT respectively, which correspond to the $90\%$ sparsity in Tab. \ref{tab:pretrain_iterations}.}
\label{tab:pretrain_time}
\end{table*}
%%%%%%%%%%%%%%%%%%%%%%%%%%%%%%%%%%%%%%%%%%%

\subsection{Comparison Between BERT and RoBERTa Subnetworks}
\label{sec:appendix_a2}
\begin{figure*}[t]
\centering
\includegraphics[width=1.0\linewidth]{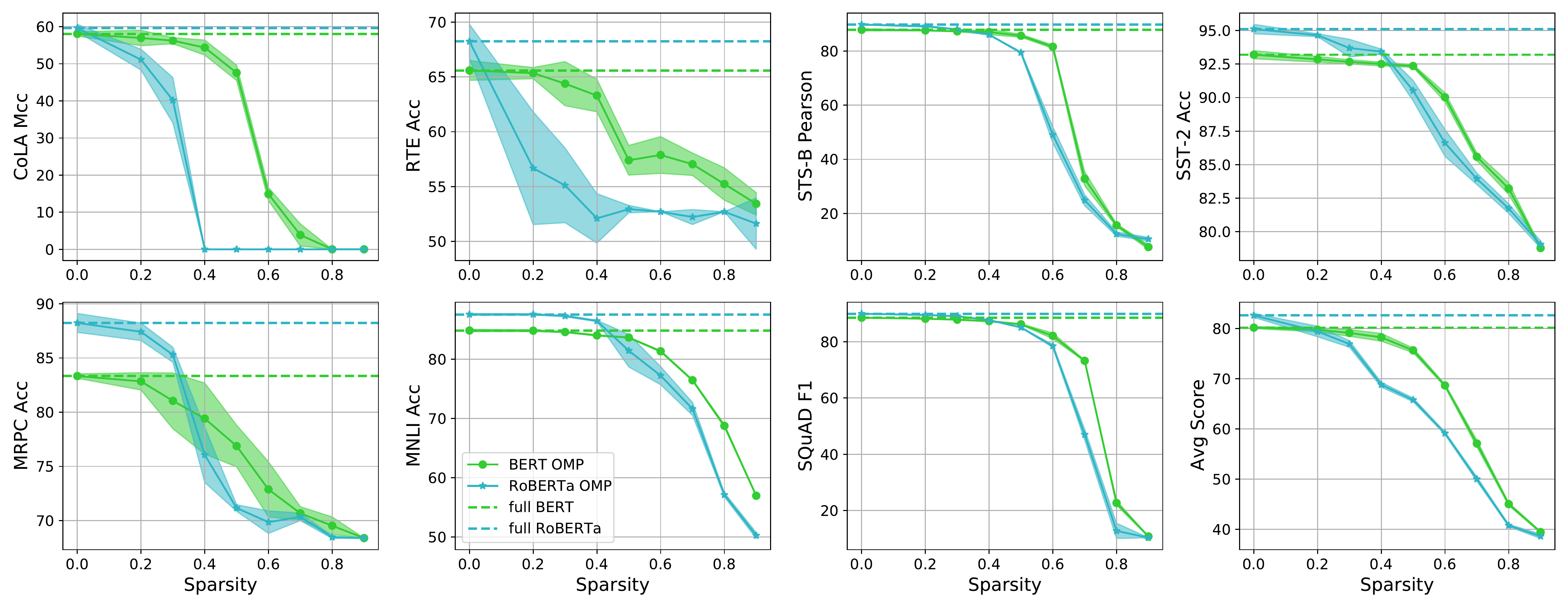}
\caption{Downstream performance of BERT and RoBERTa subnetworks found using OMP. Shadowed areas denote standard deviations.}
\label{fig:compare_bert_roberta}
\end{figure*}

In the main results of Fig. \ref{fig:main_result_bert} and Fig. \ref{fig:main_result_roberta}, we compare the fine-tuning performance of subnetworks of the same PLM but found using different methods. In this section, we give a comparison between subnetwords of $\mathrm{BERT}_{\mathrm{BASE}}$ and $\mathrm{RoBERTa}_{\mathrm{BASE}}$. As shown in Fig. \ref{fig:compare_bert_roberta}, RoBERTa consistently outperforms BERT as a full model. However, as we prune the pre-trained weights accroding to the magnitudes, the performance of RoBERTa declines more sharply than BERT, leading to worse results of RoBERTa subnetworks when crossing a certain sparsity threshold. This phenomenon suggests that, compared with BERT, RoBERTa is less robust to task-agnostic magnitude pruning. More empirical and theoretical analysis are required to understand the underlying reasons.

\subsection{Pre-training Performance and Single Task Downstream Performance}
\label{sec:appendix_a3}
\begin{figure*}[t]
\centering
\includegraphics[width=1.0\textwidth]{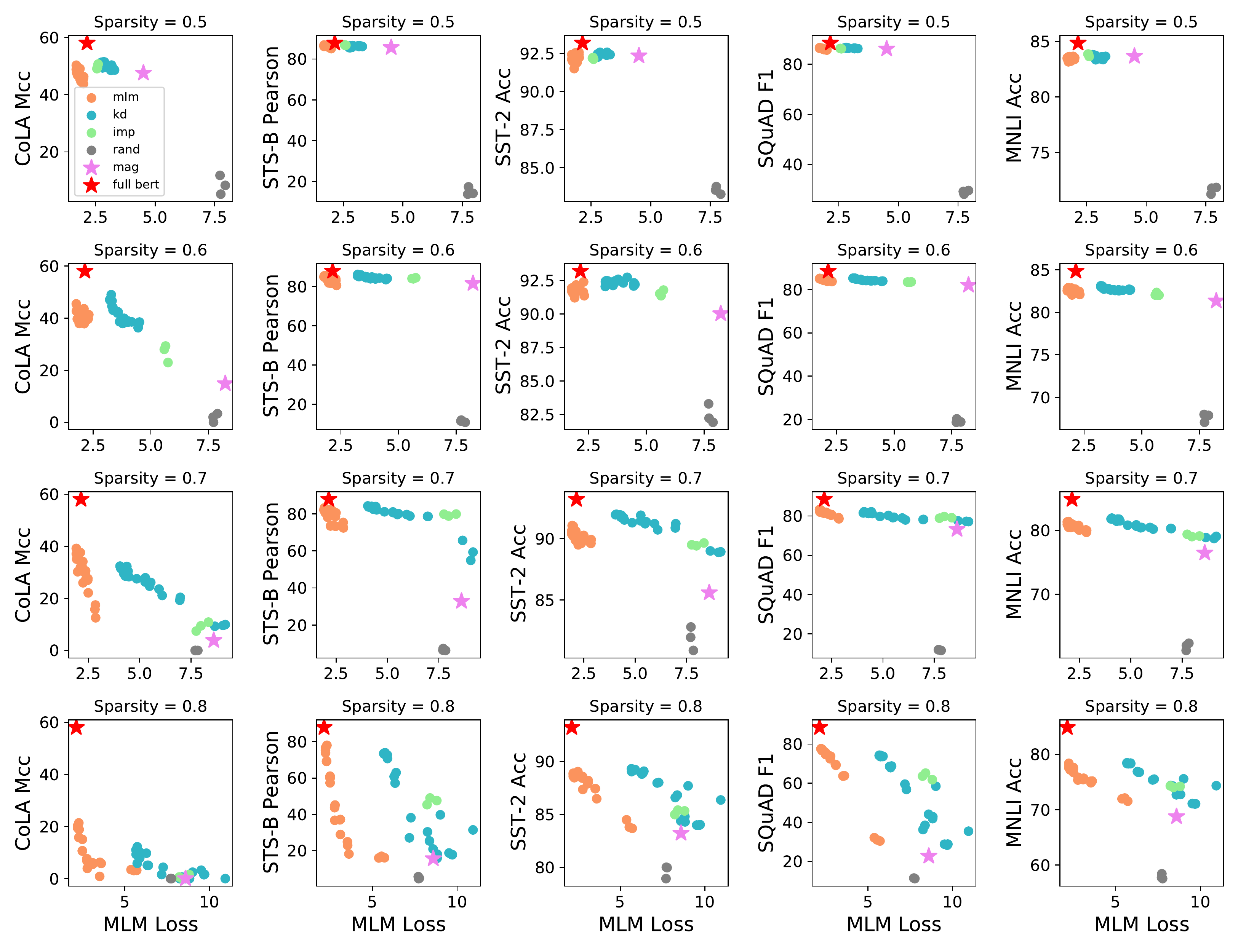}
\caption{MLM dev loss and single task downstream performance of $\mathrm{BERT}_{\mathrm{BASE}}$ subnetworks. The results of TAMT are obtained from the masks along the training process, and the results of IMP and Rand are from different seeds.}
\label{fig:mlm_loss_acc}
\end{figure*}

\begin{figure*}[t]
\centering
\includegraphics[width=1.0\textwidth]{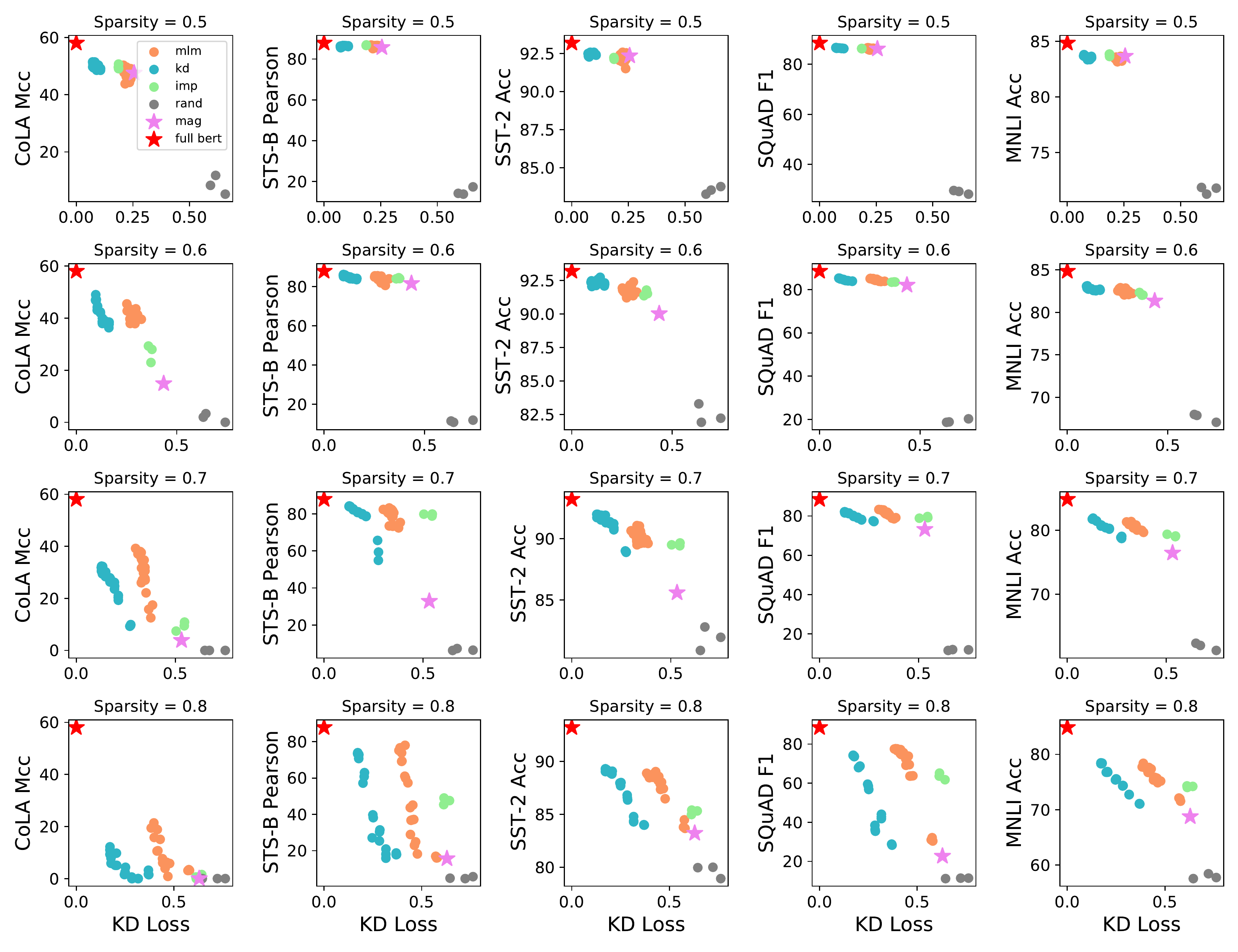}
\caption{KD dev loss and single task downstream performance of $\mathrm{BERT}_{\mathrm{BASE}}$ subnetworks. The results of TAMT are obtained from the masks along the training process, and the results of IMP and Rand are from different seeds.}
\label{fig:kd_loss_acc}
\end{figure*}

The relation between pre-training performance and overall downstream performance is illustrated in Fig. \ref{fig:mask_loss}. Here in this appendix, we provide the detailed results about each single downstream task, as shown in Fig. \ref{fig:mlm_loss_acc} and Fig. \ref{fig:kd_loss_acc}. As we can see, the pattern in each single task is general the same as we discussed in Section \ref{sec:pretrain_performance_transferability}. When the model sparsity is higher than $50\%$, TAMT promotes the performance of OMP in terms of both pre-training tasks and downstream tasks, and improves over IMP with perceivable margin. As shown in Fig. \ref{fig:main_result_bert} and Fig. \ref{fig:main_result_roberta} of the main paper, both IMP and TAMT display no obvious improvement over OMP on MRPC and RTE (but no degradation as well). Therefore, we do not report the comparison on these two datasets.

\subsection{Pre-training Iteration and Single Task Downstream Performance}
\label{sec:appendix_a4}
\begin{figure*}[t]
\centering
\includegraphics[width=1.0\textwidth]{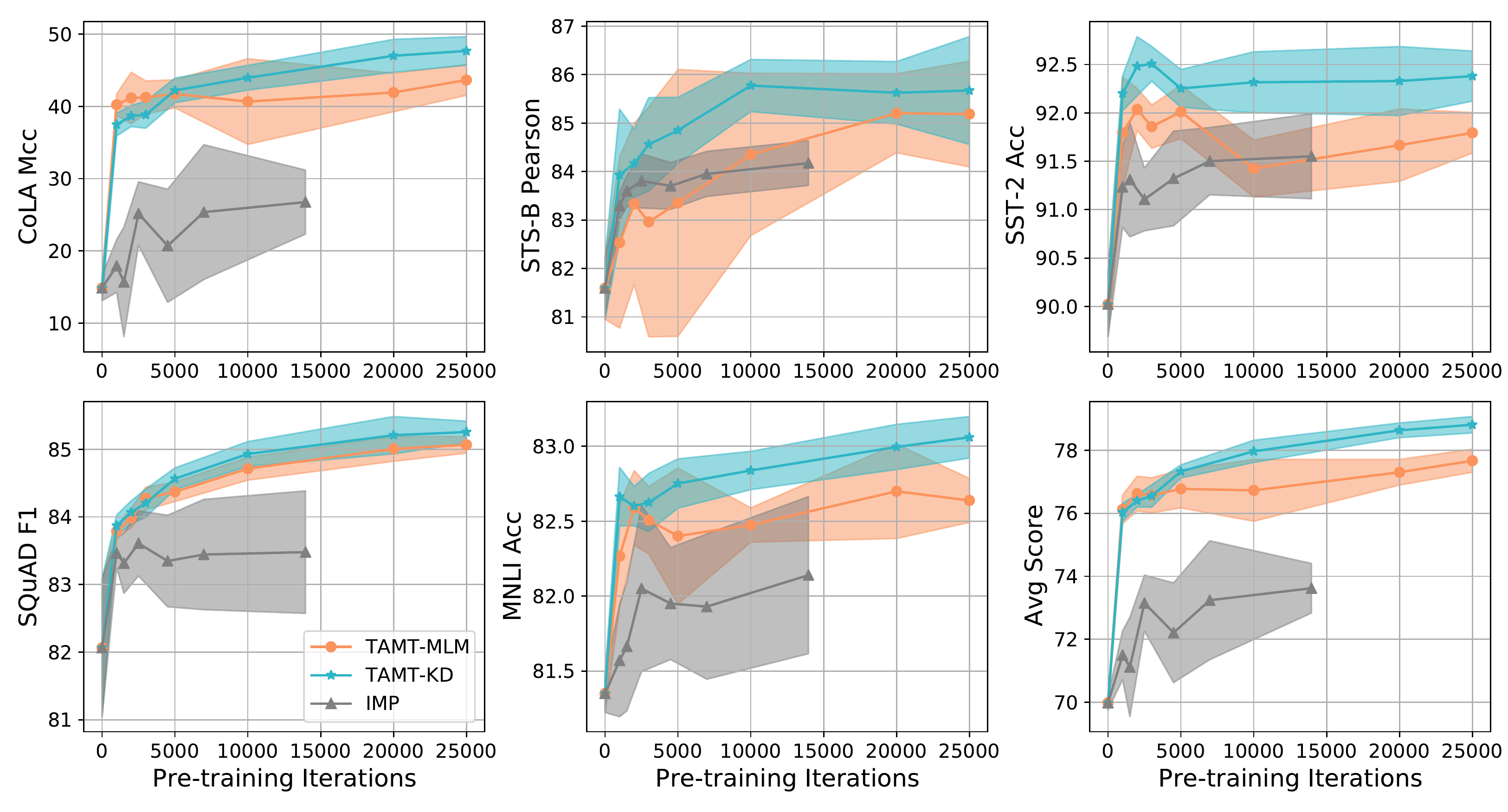}
\caption{The downstream performance of $60\%$ sparse $\mathrm{BERT}_{\mathrm{BASE}}$ subnetworks on each single task, with increased pre-training iterations.}
\label{fig:step_acc_single_task_0.6}
\end{figure*}

\begin{figure*}[t]
\centering
\includegraphics[width=1.0\textwidth]{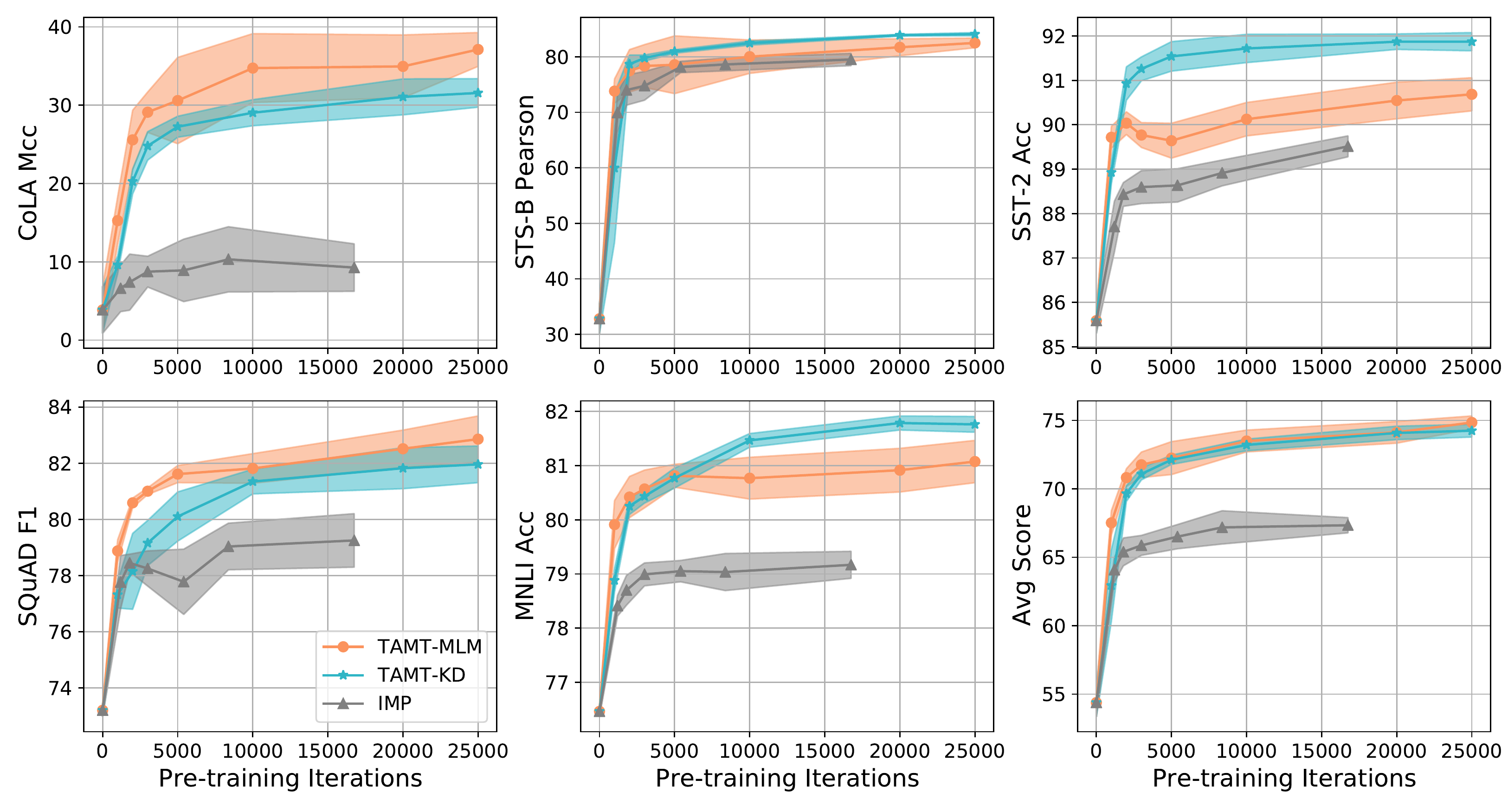}
\caption{The downstream performance of $70\%$ sparse $\mathrm{BERT}_{\mathrm{BASE}}$ subnetworks on each single task, with increased pre-training iterations.}
\label{fig:step_acc_single_task_0.7}
\end{figure*}

\begin{figure*}[t]
\centering
\includegraphics[width=1.0\textwidth]{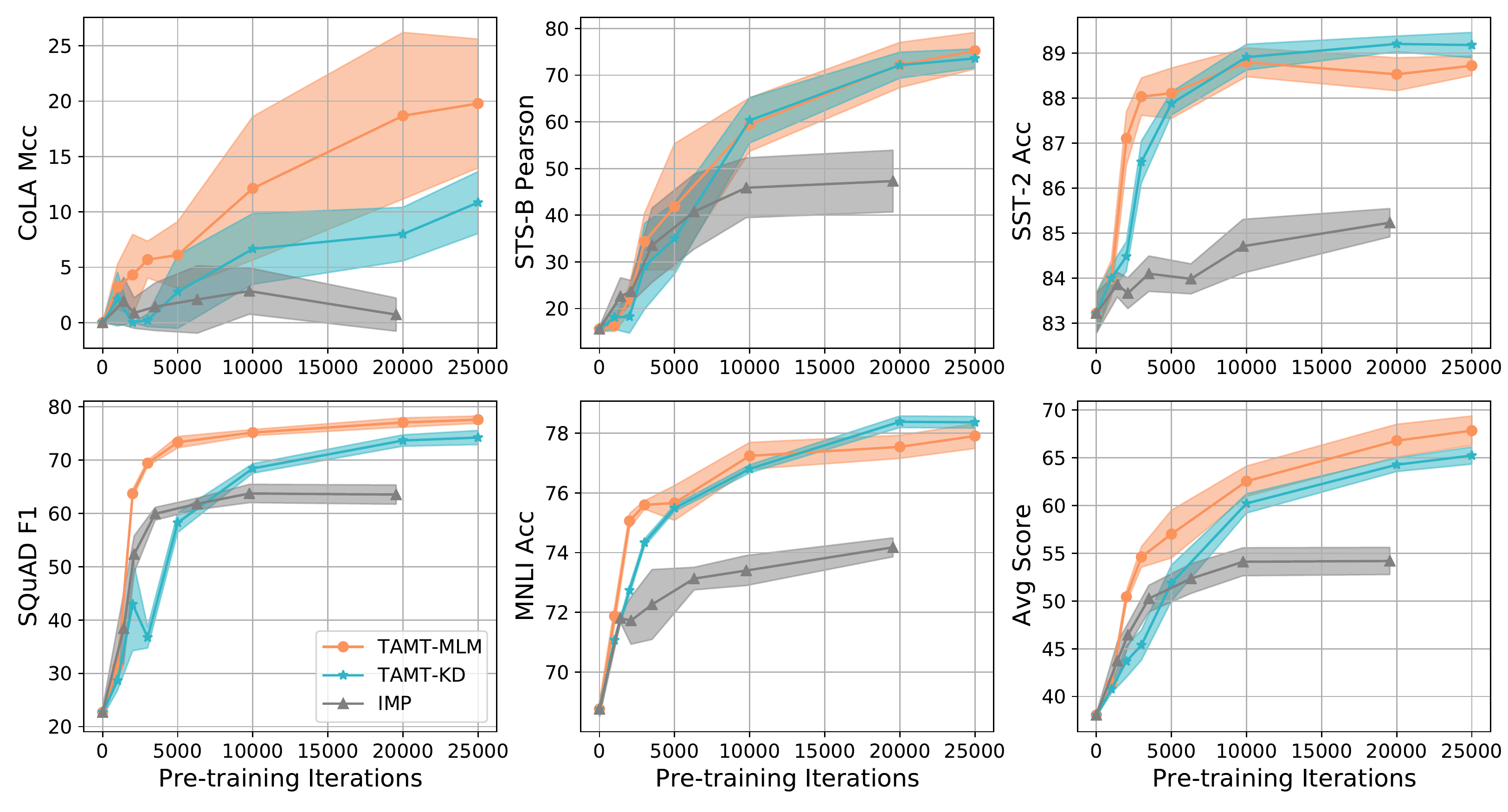}
\caption{The downstream performance of $80\%$ sparse $\mathrm{BERT}_{\mathrm{BASE}}$ subnetworks on each single task, with increased pre-training iterations.}
\label{fig:step_acc_single_task_0.8}
\end{figure*}

In Fig. \ref{fig:mask_step}, we show the overall downstream performance at $70\%$ sparsity with the increase of mask training iterations. Here, we report the results of each single downstream task from $60\% \sim 80\%$ sparsities, which are shown in Fig. \ref{fig:step_acc_single_task_0.6}, Fig. \ref{fig:step_acc_single_task_0.7} and Fig. \ref{fig:step_acc_single_task_0.8}. We can see that: 1) The single task performance of both TAMT-MLM and TAMT-KD grows faster than IMP at $60\%$ and $70\%$ sparsity, with the only exception of STS-B, where TAMT-MLM and IMP are comparable in the early stage of pre-training. 2) The MLM and KD objectives are good at different sparsity levels and different tasks. TAMT-KD performs the best at $60\%$ sparsity, surpassing TAMT-MLM on all the five tasks. In contrast, TAMT-MLM is better at higher sparsities. 3) At $80\%$ sparsity, the searching efficiency of the KD objective is not desirable, which requires more pre-training steps to outperform IMP on CoLA, STS-B, SQuAD and the overall performance. However, the advantage of TAMT-MLM is still obvious at $80\%$ sparsity.

\subsection{Subnetwork Similarity at Different Sparsities}
\label{sec:appendix_a5}
\begin{figure*}[t]
\centering
\includegraphics[width=1.0\textwidth]{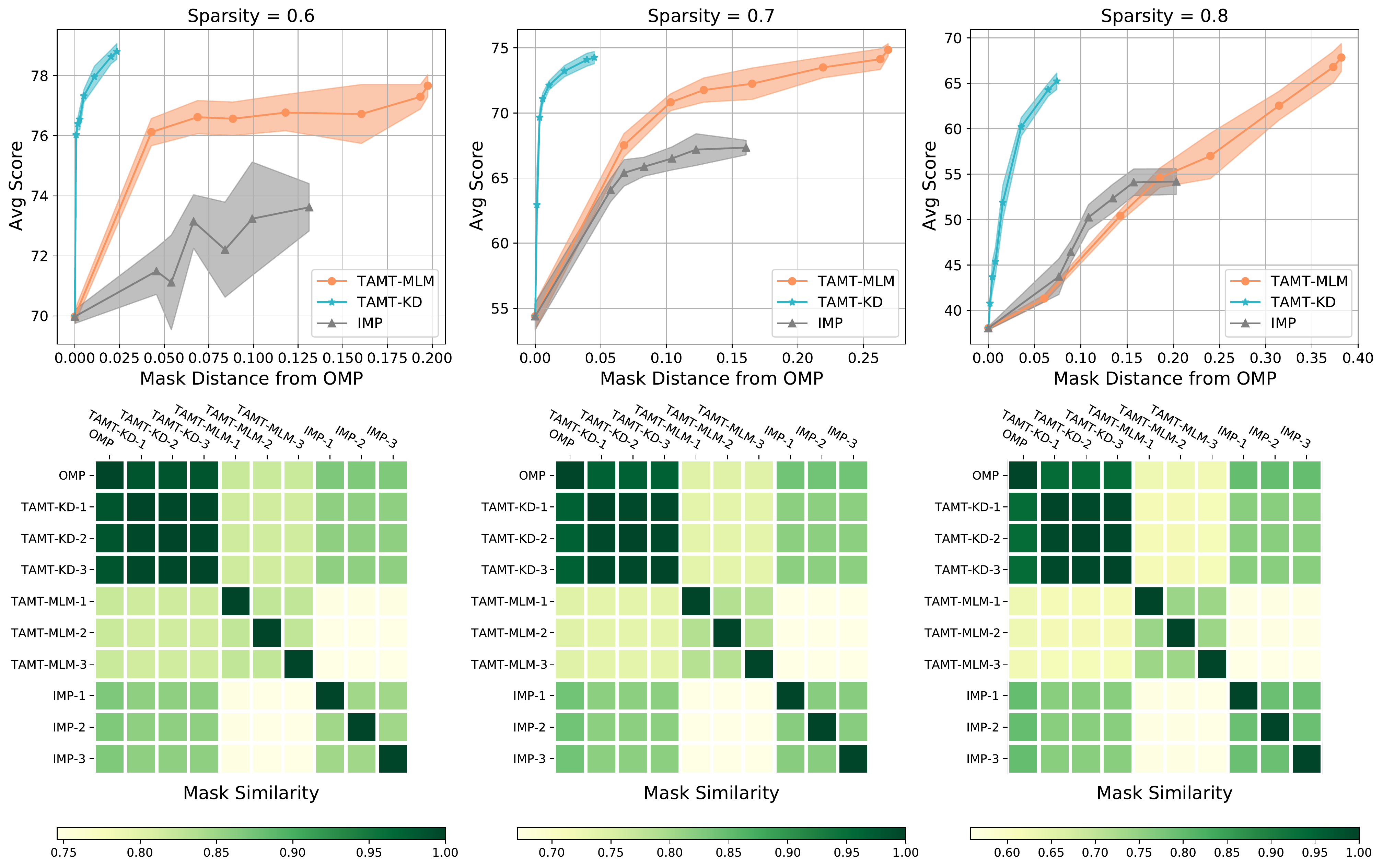}
\caption{Upper: The downstream performance of masks with varying distances from the OMP mask. Shadowed areas denote standard deviations. Lower: The similarity between masks searched using different methods. The masks are the same as those used to report the main results. The suffix numbers indicate different seeds. The masks are from $\mathrm{BERT}_{\mathrm{BASE}}$.}
\label{fig:mask_sim_more_sparsity}
\end{figure*}

In Section \ref{sec:mask_dist_sim}, we analyse the similarity between subnetworks at $70\%$ sparsity. In Fig. \ref{fig:mask_sim_more_sparsity}, we present additional results of subnetworks at different sparsities. We can see that the general pattern, as discussed in Section \ref{sec:mask_dist_sim}, is the same across $60\%$, $70\%$ and $80\%$ sparsities. However, as sparsity grows, different searching methods becomes more distinct from each other. For instance, the similarity between TAMT-MLM and IMP subnetworks decreases from 0.75 at $60\%$ sparsity to less than 0.6 at $80\%$ sparsity. This is understandable because the higher the sparsity, the lower the probability that two subnetworks will share the same weight.

\section{More Information about Experimental Setups}
\subsection{Datasets and Evaluation}
\label{sec:appendix_b1}
For pre-training, we adopt the WikiText-103 dataset \footnote{WikiText-103 is available under the Creative Commons Attribution-ShareAlike License (\url{https://en.wikipedia.org/wiki/Wikipedia:Text_of_Creative_Commons_Attribution-ShareAlike_3.0_Unported_License})} for language modeling. WikiText-103 is a collection of articles on Wikipedia and has over 100M tokens. Such data scale is relatively small for PLM pre-training. However, we find that it is sufficient for mask training and IMP to discover subnetworks with perceivable downstream improvement. 

For the downstream tasks, we use six datasets from the GLUE benchmark and the SQuAD v1.1 dataset \footnote{SQuAD is available under the CC BY-SA 4.0 license.}. The GLUE benchmark is intended to train, evaluate, and analyze NLU systems. Our experiments include the tasks of CoLA for linguistic acceptability, SST-2 for sentiment analysis, RTE and MNLI for natural language inference, MRPC and STS-B for semantic matching/similarity. The SQuAD dataset is for the task of question answering. It consists of questions posed by crowdworkers on a set of Wikipedia articles. Tab. \ref{tab:setup} summarizes the dataset statistics and evaluation metrics. All the datasets are in English language.

\subsection{Implementation Details}
\label{sec:appendix_b2}
The hyper-parameters for pre-training and fine-tuning are shown in Tab. \ref{tab:setup}. The pre-training setups of IMP basically follow \cite{BERT-LT}, except for the number of training epochs, because we use different pre-training datasets. Since we aim at finding universal PLM subnetworks that are agnostic to the downstream tasks, we do \textbf{not} perform hyper-parameter search for TAMT based on the downstream performance. The pre-training hyper-parameters in Tab. \ref{tab:setup} are determined as they can guarantee stable convergence on the pre-training tasks.

For fair comparison between TAMT and IMP, we control the number of pre-training iterations (i.e., the number of gradient descent steps) to be the same. Considering that the IMP subnetworks of different sparsities are obtained from different pre-training iterations, we adjust the pre-training iterations of TAMT accordingly. Specifically, we set the maximum number of pre-training epochs to 2 for IMP, which equals to 27.92K training iterations. Thus, the sparsity is increased by 10\% every 2.792K iterations. Tab. \ref{tab:pretrain_iterations} shows the number of pre-training iterations for IMP and TAMT subnetworks at $20\% \sim 90\%$ sparsity. Note that the final training iteration does not equal to 27.92K at $100\%$ sparsity according to Tab. \ref{tab:pretrain_iterations}. This is because we prune to $10\%$ sparsity at the $0^{th}$ iteration, which follows the implementation of \citet{BERT-LT}.

The hyper-parameters for downstream fine-tuning follow the standard setups of \cite{huggingface,BERT-LT}. We use the same set of hyper-parameters for all the subnetworks, as well as the full models. We perform evaluations during the fine-tuning process, and the best result is reported as the downstream performance.

Training and evaluation are implemented on Nvidia V100 GPU. The codes are based on the Pytorch framework\footnote{https://pytorch.org/} and the huggingface \textit{Transformers} library\footnote{https://github.com/huggingface/transformers} \cite{huggingface}. Tab. \ref{tab:pretrain_time} shows the pre-training time of IMP and TAMT.

%%%%%%%%%%%%%%%%%%%%%%%%%%%%%%%%%%%%%%%%%%%%%%
%%%%%%%%%%%%%%%%%%%%%%%%%%%%%%%%%%%%%%%%%%%%%%
%%%%%%%%%%%%%%%%%%%%%%%%%%%%%%%%%%%%%%%%%%%%%%
%%%%%%%%%%%%%%%%%%%%%%%%%%%%%%%%%%%%%%%%%%%%%%
%%%%%%%%%%%%%%%%%%%%%%%%%%%%%%%%%%%%%%%%%%%%%%
%%%%%%%%%%%%%%%%%%%%%%%%%%%%%%%%%%%%%%%%%%%%%%
%%%%%%%%%%%%%%%%%%%%%%%%%%%%%%%%%%%%%%%%%%%%%%
%%%%%%%%%%%%%%%%%%%%%%%%%%%%%%%%%%%%%%%%%%%%%%
%%%%%%%%%%%%%%%%%%%%%%%%%%%%%%%%%%%%%%%%%%%%%%
%%%%%%%%%%%%%%%%%%%%%%%%%%%%%%%%%%%%%%%%%%%%%%
%%%%%%%%%%%%%%%%%%%%%%%%%%%%%%%%%%%%%%%%%%%%%%
%%%%%%%%%%%%%%%%%%%%%%%%%%%%%%%%%%%%%%%%%%%%%%

%%%%%%%%%%%%%%%%%%%%%%%%%%%%%%%%%%%%%%%%%%%%%%
%%%%%%%%%%%%%%%%%%%%%%%%%%%%%%%%%%%%%%%%%%%%%%

\end{document}